\documentclass[journal]{IEEEtran}
\usepackage{graphicx}
\usepackage{epstopdf} 
\usepackage{xcolor}
\usepackage{amsmath}
\usepackage{amsfonts}
\usepackage{fixltx2e}
\usepackage{hyperref}
\usepackage{subcaption}

\usepackage{multirow}

\usepackage{ifpdf}

\ifCLASSINFOpdf
\else
\fi
\begin{document}
%
\title{A Probabilistic framework for Quantum Clustering}
%
%
%

\author{Ra{\'u}l V. Casa\~na-Eslava,
        Paulo J. G. Lisboa,
        Sandra Ortega-Martorell,
        Ian H. Jarman
        and~Jos{\'e} D. Mart{\'i}n-Guerrero
\thanks{R. V. Casa\~na-Eslava, P. J. G. Lisboa, Sandra Ortega-Martorell and I. H. Jarman are with the Department of Applied Mathematics
       Liverpool John Moores University (LJMU)
       3 Byrom Street, Liverpool, L3 3AF, UK}
\thanks{J. D. Mart{\'i}n-Guerrero is with Departament d'Enginyeria Electr\`onica - ETSE
       Universitat de Val\`{e}ncia (UV)
       Av. Universitat, SN, 46100 Burjassot, Val\`{e}ncia, Spain}
}

\maketitle

\begin{abstract}
Quantum Clustering is a powerful method to detect clusters in data with mixed density. However, it is very sensitive to a length parameter that is inherent to the Schr\"odinger equation. In addition, linking data points into clusters requires local estimates of covariance that are also controlled by length parameters. This raises the question of how to adjust the control parameters of the Schr\"odinger equation for optimal clustering.
We propose a probabilistic framework that provides an objective function for the goodness-of-fit to the data, enabling the control parameters to be optimised within a Bayesian framework. This naturally yields probabilities of cluster membership and data partitions with specific numbers of clusters. The proposed framework is tested on real and synthetic data sets, assessing its validity by measuring concordance with known data structure by means of the Jaccard score (JS). This work also proposes an objective way to measure performance in unsupervised learning that correlates very well with JS.
\end{abstract}


%
\IEEEpeerreviewmaketitle

\section{Introduction}
\label{Intro}
%
%
%
%
%
%

\IEEEPARstart{Q}{uantum} Clustering (QC) is an appealing paradigm inspired by the Schr\"odinger equation~\cite{NIPS2001_2083} with potential to identify and track connected regions while separating them from other nearby clusters.
However, the method would benefit from a stronger theoretical basis to assess the goodness of fit to the density distribution of the data, in particular to guide the choice of control parameters which determine the number of clusters detected. In addition, QC is prone to fragment the data into many small clusters that may comprise outliers, without clearly defined means to control this process. 
	We propose a probabilistic framework to address these issues. This framework is applied to an extension to QC using local estimates of the central length scale parameter, which enables the model to accurately detect clusters with very different data densities.

\medskip

The starting point is the original quantum clustering algorithm introduced by~\cite{NIPS2001_2083} which generates a potential function $V(\mathbf{x})$ from a wave function $\Psi(\mathbf{x})$ as a constant energy solution of the time-independent Schr\"odinger equation:

\begin{eqnarray}
\label{eq:shcrodinger_eq}
H \Psi \equiv  \left(-\frac{\sigma^2}{2} \nabla^2+V(\mathbf{x})\right)\Psi(\mathbf{x})=E\Psi(\mathbf{x})
\end{eqnarray}
\noindent where $H$ is the Hamiltonian and $E$ the constant total energy.

In the original formulation the wave function represented a Parzen estimator with a given length scale parameter, $\sigma$.
Given a potential function generated from the wave function using the Schr\"odinger equation, the allocation of individual data points to clusters was determined by the use of gradient descent~\cite{NIPS2001_2083} to find local minima and allocate clusters based on maximum probability of cluster membership, although local Hessian modes in a potential lattice have also been used for this purpose~\cite{ar:nasios2006}.

While the wave function providing data density estimates need not be Gaussian e.g., B-splines~\cite{gelman2014bayesian}, Vector Quantization~\cite{gray1984vector} or the Epanechnikov kernel~\cite{silverman1986density,comaniciu1999distribution} with optimal efficiency, exponential distributions are generally preferred due to their smoothness since the wave function has to be differentiable up to third order in the potential function:

\begin{eqnarray}
\label{eq:potential}
V(\mathbf{x})= E + \frac{\sigma^2}{2} \frac{\nabla^2 \Psi(\mathbf{x})}{\Psi(\mathbf{x})}
\end{eqnarray}

We propose to define the wave function by assigning a normalized Gaussian function to each observation in the training set, centred on the observed data point and with the covariance matrix estimated locally from a set number of nearest neighbours (NNs) and assumed to be diagonal.
%
%
%

This is in contrast with Mixtures of Gaussian Models (MGMs) where the mean and covariance parameters are not tied to data points and are fitted using Maximization-Expectation~\cite{friedman2001elements}. Our approach is equivalent to a generative model with a kernel representing inherent noise and a prior that is uniformly distributed across all of the observations. The total wave function thus comprises many narrow Gaussians creating an aggregate density function that links neighbouring data points to generate a smooth and connected valley in the potential function.

Clearly the length scale of the exponential functions, $\sigma$, which parameterises the variance of the noise estimate, is of critical importance since it determines the overlap between the wave function components from neighbouring observations and so has a critical impact on the shape and smoothness of the resulting potential function, by affecting the number of local minima and, consequently, also the number of clusters.
 
A recent publication~\cite{cui2018development} recognises the difficulty in mapping local data structure with local potential functions and proposes training a self-organised neural network with radial basis functions as the basic computational units. This empirical approach is effective although it does not claim to optimise the model parameters, because of the complex structure of local minima of the potential surface. 

Regarding metric learning,Topological Data Analysis (TDA)~\cite{wasserman2018topological,bubenik2015statistical} can be considered a similar approach in terms of data-structure characterization through the search of distance-parameter stabilization, where the clustering is based on this topology persistence~\cite{chazal2013persistence}. However, TDA lacks the methodical and objective criteria for selecting the hierarchical level of the dendrograms. On the contrary, we propose a Bayesian interpretation of our generative model, in order to infer probabilistic measures for the goodness-of-fit of the data, providing a score function for parameter selection. The probabilistic QC (PQC) outperforms TDA in terms of Jaccard scores (JS), as it will be shown in Section~\ref{Res}.

It was the dependence of the original QC on the band-width selection of the Parzen window which originally led to the use of k-nearest neighbours (KNN) in kernel estimators of the local sample covariance~\cite{ar:nasios2006,inp:nasios2005}. Unfortunately, the efficiency of KNN estimators varies considerably depending on the structure of the data~\cite{casana_qc_nc_2017}. An alternative approach is considered in~\cite{zelnik2005self,li2016quantum}, where the kernel scale is locally estimated. In~\cite{vincent2003manifold}, a probability density function is estimated using a manifold Parzen window, rendering the Gaussian function non-spherical. Summing up, the determination of a suitable kernel length to discriminate clusters from the QC potential lacked a defined framework to measure goodness of fit to the data, making it difficult to optimise this critical parameter.

We propose a probabilistic interpretation of quantum clustering through the use of wave functions comprising normalised joint probability distributions. This enables the length parameters for local covariance estimation to be optimised by maximising a Bayesian probability of cluster allocation. 
An empirical evaluation with synthetic and real-world data shows that the approach is robust for clustering complex data by maximising the probability of cluster membership without prior knowledge of the correct number of clusters. After obtaining the probability of cluster membership, the standard Bayesian framework can be used to detect outliers. 

Furthermore, a positive definite likelihood function of cluster membership can be optimised to select the bandwidth of the Gaussian functions, namely the number of KNNs used for local covariance estimation, which is the only free parameter in the model. This underlines PQC as a plausible method for the detection of hierarchical data structure.


The proposed framework addresses two complementary challenges for current methods. Regular mixtures of Gaussians cannot resolve the number of clusters, hence requiring a preset value of K. While this is addressed in part by DBSCAN (Density-based spatial clustering of applications with noise)~\cite{ar:DBSCAN}, it still remains difficult to estimate the hyper-parameter corresponding to the minimum number of core points when the data are heavily heteroscedastic. In this paper, we propose and demonstrate a framework to select efficient parameters to map, with quantum clustering, complex data structure with no prior knowledge of the number of clusters.

The rest of the paper is structured as follows: Section~\ref{Methods} introduces the original QC in Subsection~\ref{origQC}, then the proposed QC improvements based on KNNs are described in Subsection~\ref{QC2}, the manifold QC in \ref{QC3}, and PQC in~\ref{PQC}; subsection~\ref{ANLL} shows how the likelihood function based on $P(K|\mathbf{X})$ becomes an effective and objective way of assessing the performance of unsupervised classification, becoming a major contribution of this paper. Section~\ref{data} presents the data sets used to test the performance of the proposed clustering method with results reported in Section~\ref{Res}. Section~\ref{Conc} concludes with a critical summary of PQC, the conclusions that can be drawn from the work and it also identifies directions for further work. The paper also includes two appendices, one devoted to the description of the cluster-allocation process, and the other one about the selection of a local-covariance threshold.

\section{Methods}
\label{Methods}

In case of heterogeneous features, the data are sphered by standardizing each dimension to the $z$-score, in order to provide a uniform length scale across all dimensions. Additionally, the data is scaled by a constant, $1/\lambda$, to make the length scale uniform when the stochastic gradient descent (SGD) is applied: 

\begin{eqnarray}
\label{eq:scale}
\lambda = \frac{\sum_{i=1}^{n} \|\mathbf{x}_i \|}{n}
\end{eqnarray}

\noindent where $n$ is the length of vector $\mathbf{x}$, and $\|\mathbf{x}\|$ is the $L^2$ norm.

\subsection{\texorpdfstring{Original Quantum Clustering, $QC_\sigma$}{}}
\label{origQC}

${QC}_{\sigma}$ starts by defining a wave function as a Parzen estimator, from which a convex potential function is derived by the Schr\"odinger equation. Cluster allocation consists in identifying which regions belong to each local minimum of the potential function, originally through a gradient descent (GD).

Gaussian kernels associated with each observation add together to make the wave function~(\ref{eq:wave1}):

\begin{eqnarray}
\label{eq:wave1}
\Psi(\mathbf{x})= \sum_{i=1}^{n} \psi_i\left(\mathbf{x}\right) = \sum_{i=1}^{n} e^{-\frac{\left(\mathbf{x}-\mathbf{x}_i\right)^2}{2\sigma^2}}
\end{eqnarray}

\noindent where $n$ is the sample size and $\sigma$ a global length scale comprising a single hyper-parameter to adjust. The Gaussian normalisation $(\sqrt{2\pi}\sigma)^{-d}$ is redundant as it will cancel out in the calculation of the potential function.
Applying equation~(\ref{eq:potential}):

\begin{equation}
\label{eq:V}
\begin{split}
V(\mathbf{x})= E + \frac{\sigma^2}{2} \frac{\nabla^2 \Psi(\mathbf{x})}{\Psi(\mathbf{x})} = \\
E - \frac{d}{2} + \frac{\sum_{i=1}^{n} \left(\mathbf{x}-\mathbf{x}_i\right)^2 e^{-\frac{\left(\mathbf{x}-\mathbf{x}_i\right)^2}{2\sigma^2}}}{2 \sigma^2 \Psi(\mathbf{x})}
\end{split}
\end{equation}

\noindent where $d$ is the dimension of the input space, and $E$ in this context plays the role of an offset of $V(\mathbf{x})$. If we impose that $V(\mathbf{x})\geq 0$, then $E= - min \frac{\sigma^2}{2} \frac{\nabla^2 \Psi(\mathbf{x})}{\Psi(\mathbf{x})}$. For the purposes of cluster allocation with the gradient descent, the offset values of $V(\mathbf{x})$ are irrelevant.

Therefore, ${QC}_{\sigma}$ potential is a weighted averaged function over $\Psi$, that could be expressed as the expected value of $F$ over $\Psi$, being $F_i = \left(\mathbf{x}-\mathbf{x}_i\right)^2$:

\begin{eqnarray}
\label{eq:expectedval}
{\langle F_i \rangle}_\Psi \equiv \frac {\sum_{i} F_i \psi_i}{\sum_{i} \psi_i}
\end{eqnarray}

Applying the gradient to ${\langle F \rangle}_\Psi$:

\begin{equation}
\label{eq:gradexpectedval}
\begin{split}
\nabla {\langle F_i \rangle}_\Psi = \\
{\Big \langle \nabla F_i - \frac{F_i}{\sigma^2} \left(\mathbf{x}-\mathbf{x}_i\right) \Big \rangle}_\Psi + {\Big \langle F_i \Big \rangle}_\Psi {\Big \langle \frac{\left(\mathbf{x}-\mathbf{x}_i\right)}{\sigma^2} \Big \rangle}_\Psi
\end{split}
\end{equation}

With this notation the $V(\mathbf{x})$ and $\nabla V(\mathbf{x})$ are simplified to:

\begin{eqnarray}
\label{eq:V2}
V(\mathbf{x})= E + \frac{\sigma^2}{2} \frac{\nabla^2 \Psi(\mathbf{x})}{\Psi(\mathbf{x})} = E - \frac{d}{2} +{\Big \langle \frac{\left(\mathbf{x}-\mathbf{x}_i\right)^2}{2 \sigma^2} \Big \rangle}_\Psi
\end{eqnarray}

\begin{equation}\label{eq:gradV2}
\begin{split}
\nabla V(\mathbf{x}) = 
 {\Bigg \langle \frac{\left(\mathbf{x}-\mathbf{x}_i\right)}{\sigma^2} \Bigg \rangle}_\Psi \left(1 + \bigg \langle {\frac{\left(\mathbf{x}-\mathbf{x}_i\right)^2}{2\sigma^2} \bigg \rangle}_\Psi  \right) - \\
{\Bigg \langle \frac{\left(\mathbf{x}-\mathbf{x}_i\right) \left(\mathbf{x}-\mathbf{x}_i\right)^2}{2 \sigma^4}  \Bigg \rangle}_\Psi 
\end{split}
\end{equation}

The next step in the ${QC}_{\sigma}$ is to apply the GD to allocate the clusters. Defining $\mathbf{y}_i(0) = \mathbf{x}_i$, the usual GD is:

\begin{eqnarray}
\label{eq:SGD}
\mathbf{y}_i(t+\Delta t) = \mathbf{y}_i(t) - \eta (t) \nabla V(\mathbf{y}_i(t))
\end{eqnarray}

\noindent where $\eta (t)$ is the learning rate.

We apply ADAM, a variant of SGD with an adaptive momentum term~\cite{kinga2015method} which makes it suitable for sparse gradients that commonly occur with sparse data or outliers. In order to ensure the convergence of SGD, two criteria are imposed:

\begin{eqnarray}
\label{eq:SGDconv}
max(|\Delta \mathbf{y}_i|) \leq \epsilon_y \quad \quad \quad \quad max(\Delta V(\mathbf{y}_i)) \leq \epsilon_V
\end{eqnarray}

\noindent where $\epsilon$ is the threshold\footnote{Empirically, a good value for both thresholds is $\epsilon \approx 0.001$ if the data have been rescaled to have an average length of 1.}. The first stopping criterion ensures that the updating distances in SGD are smaller than a given threshold, while the second limits the size of potential differences. The next step is to identify the clusters allocated to particular local minima of the potential function. This is detailed in Appendix~\ref{app:c_alloc}.\par

\medskip

One of the main limitations of ${QC}_{\sigma}$ is having a single length scale which sets equal width for all Gaussian functions irrespective of local density. The smaller the value of the length scale, the higher the number of detected clusters; extreme small values may lead to detect one cluster per observation while large values may find only a single cluster for all the data set. A parametrisation of the length scale that reflects variations in local density is the average of pairwise distances between observations ordered by proximity i.e., KNN, as shown below:

\begin{eqnarray}
\label{eq:sigma}
\sigma_{k\%} = \frac{1}{n}\sum_{i}^{n} \sum_{j \: \in \: knn} dist(\mathbf{x}_j,\mathbf{x}_i)
\end{eqnarray}

In order to compare different QC methodologies, we now introduce artificial data set \#1, which is further detailed in Section~\ref{art1}. It consists of four two-dimensional clusters some of which are strongly anisotropic, as well as a high-density cluster nested within a low-density one. Each cluster has 100 observations.

\medskip

\begin{figure*}
\centering
\begin{subfigure}{.5\textwidth}
  \centering
  \includegraphics[width=0.9\linewidth]{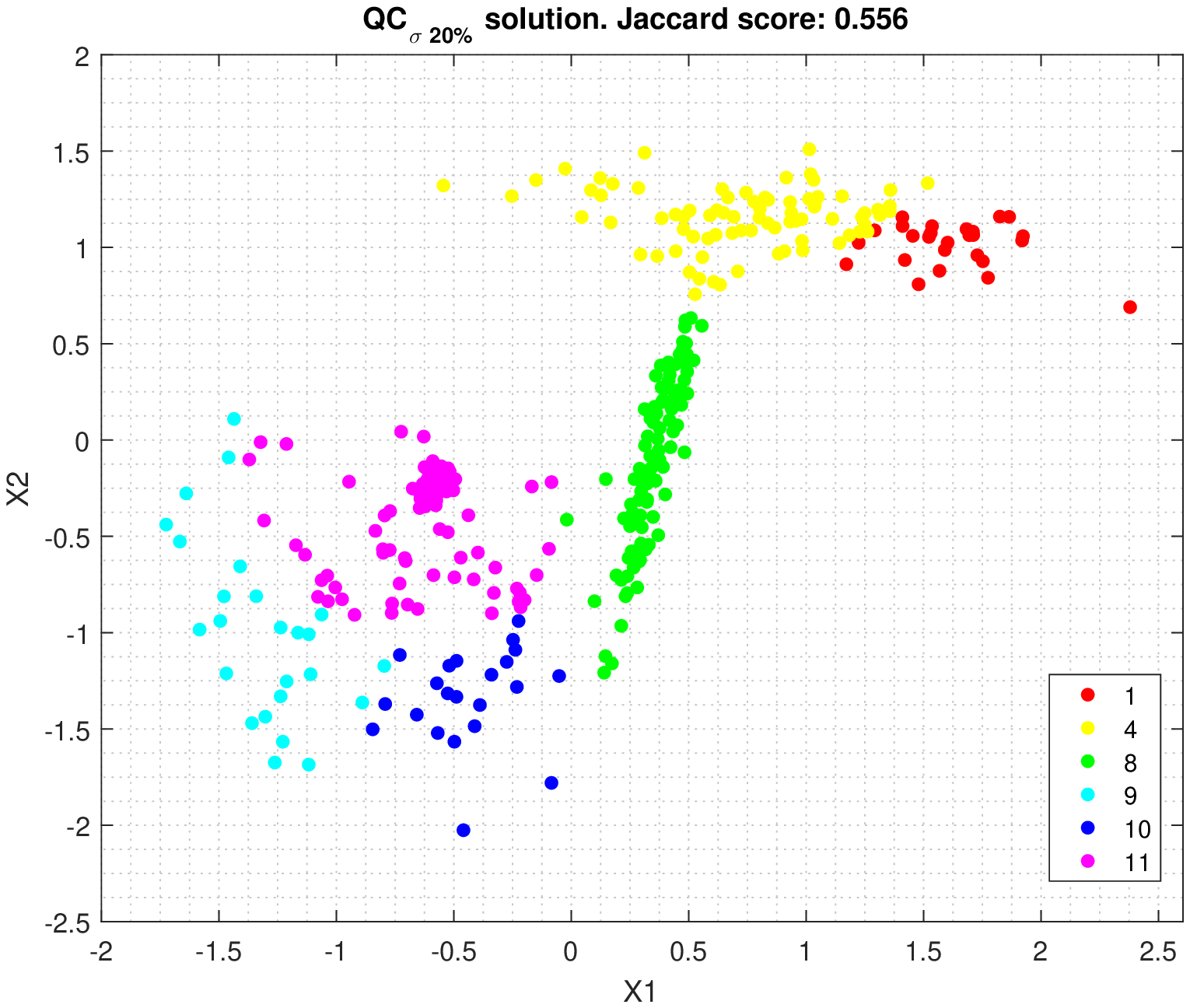}
  \caption{SGD cluster allocation $QC_{\sigma \, 20\%}$}
  \label{fig:qc1_sol}
\end{subfigure}%
\begin{subfigure}{.5\textwidth}
  \centering
  \includegraphics[width=0.9\linewidth]{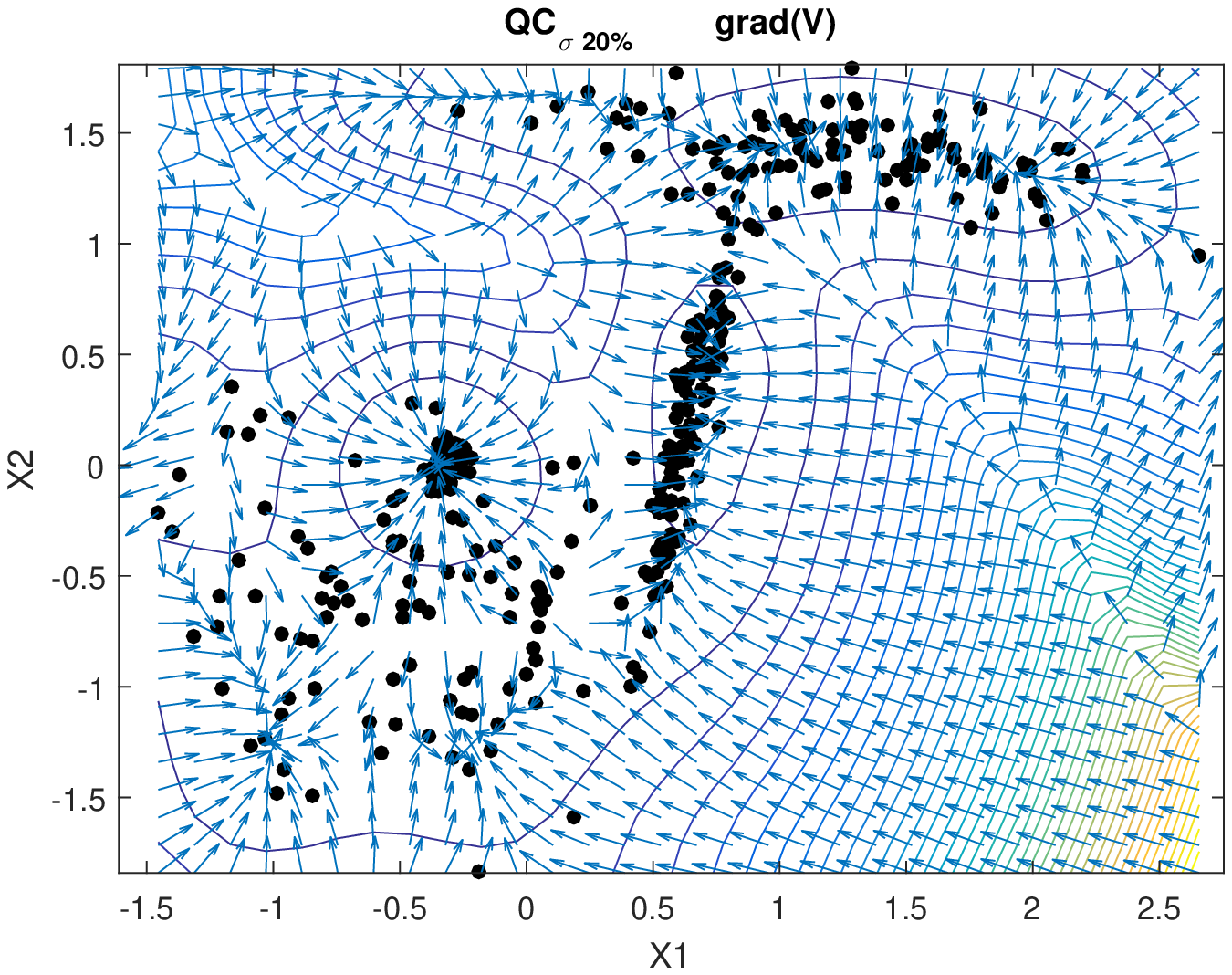}
  \caption{Gradient $QC_{\sigma \, 20\%}$}
  \label{fig:qc1_grad}
\end{subfigure}
\begin{subfigure}{.5\textwidth}
  \centering
  \includegraphics[width=0.9\linewidth]{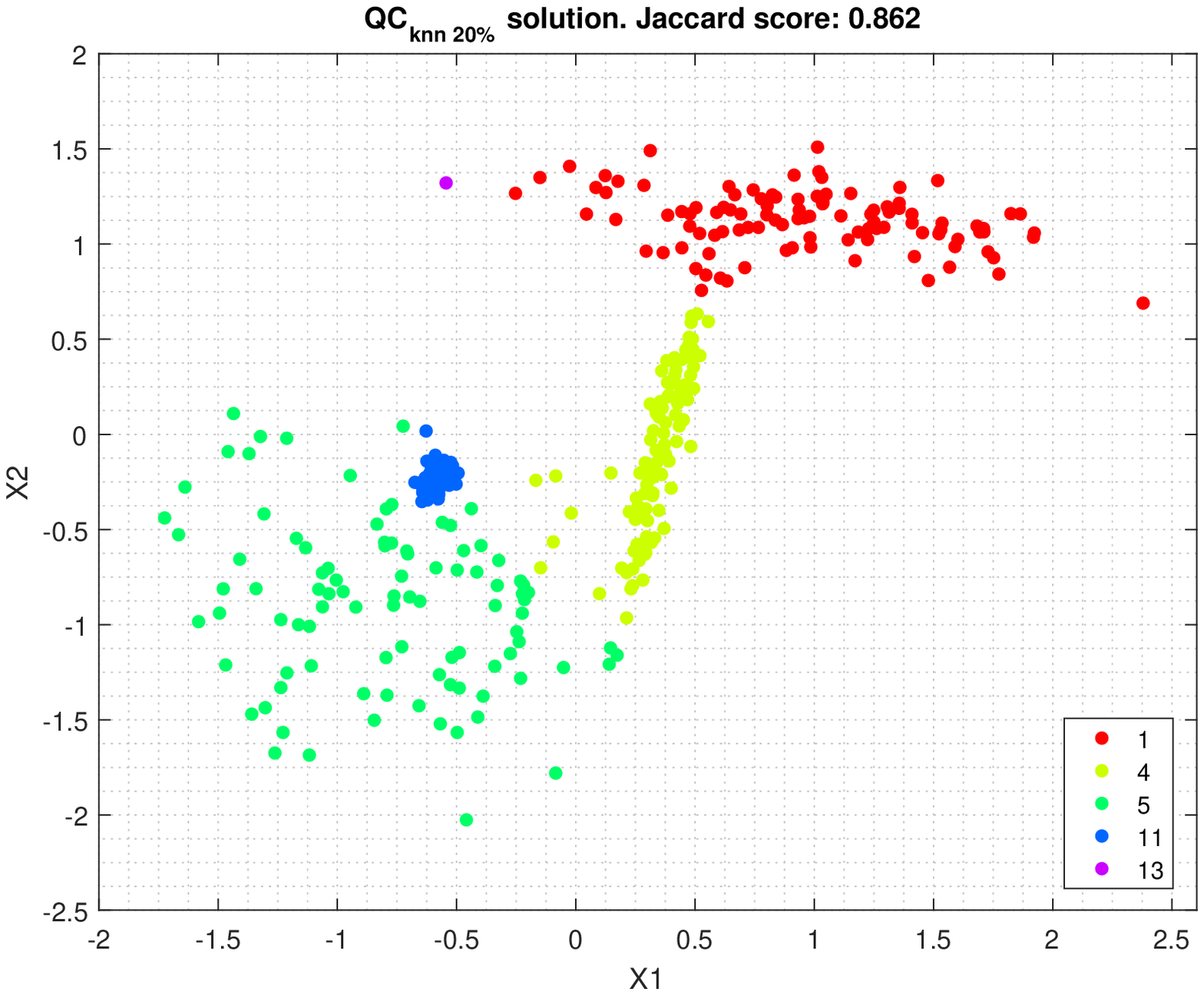}
  \caption{SGD cluster allocation $QC_{knn \, 20\%}$}
  \label{fig:qc2_sol}
\end{subfigure}%
\begin{subfigure}{.5\textwidth}
  \centering
  \includegraphics[width=0.9\linewidth]{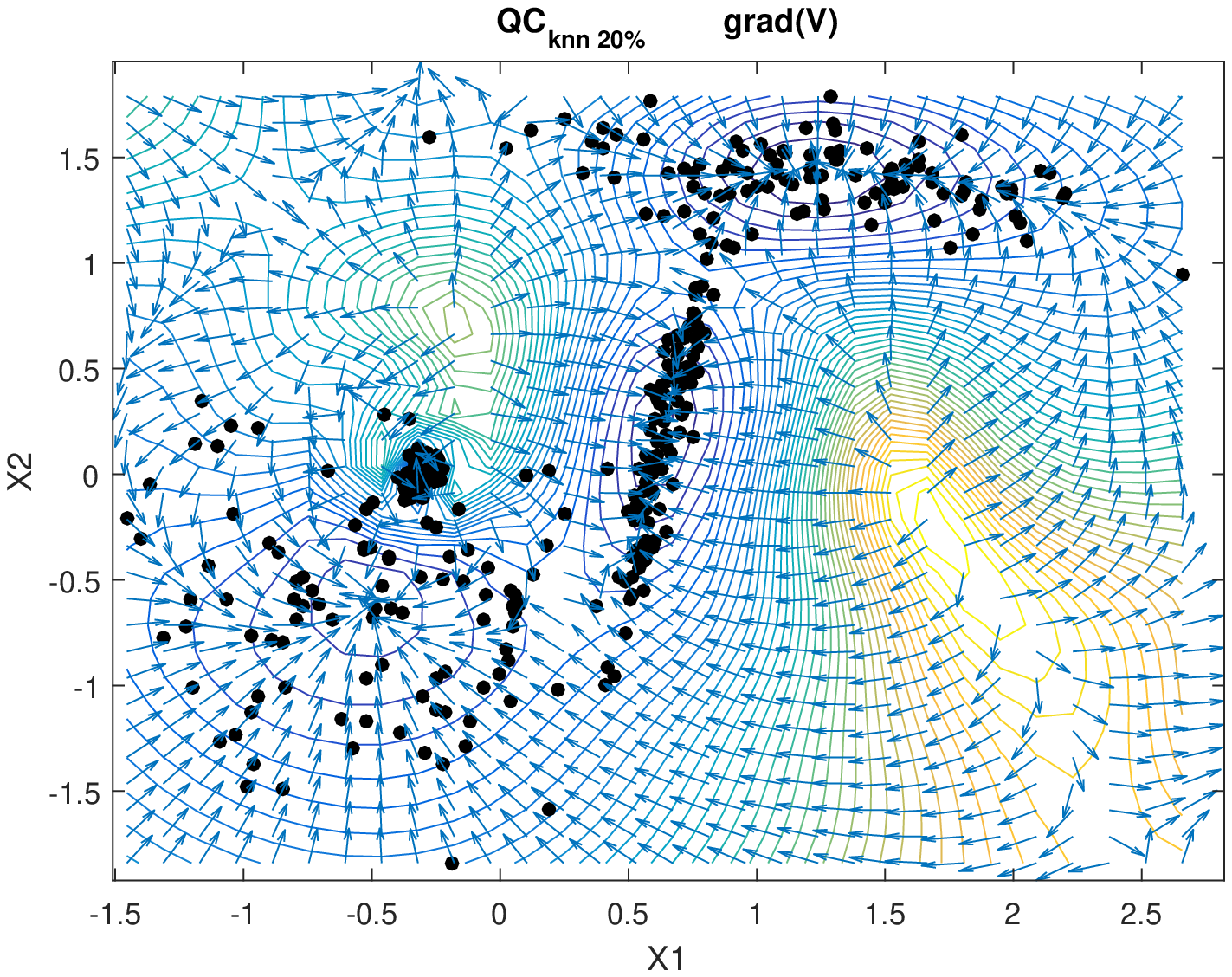}
  \caption{Gradient $QC_{knn \, 20\%}$}
  \label{fig:qc2_grad}
\end{subfigure}
\begin{subfigure}{.5\textwidth}
  \centering
  \includegraphics[width=0.9\linewidth]{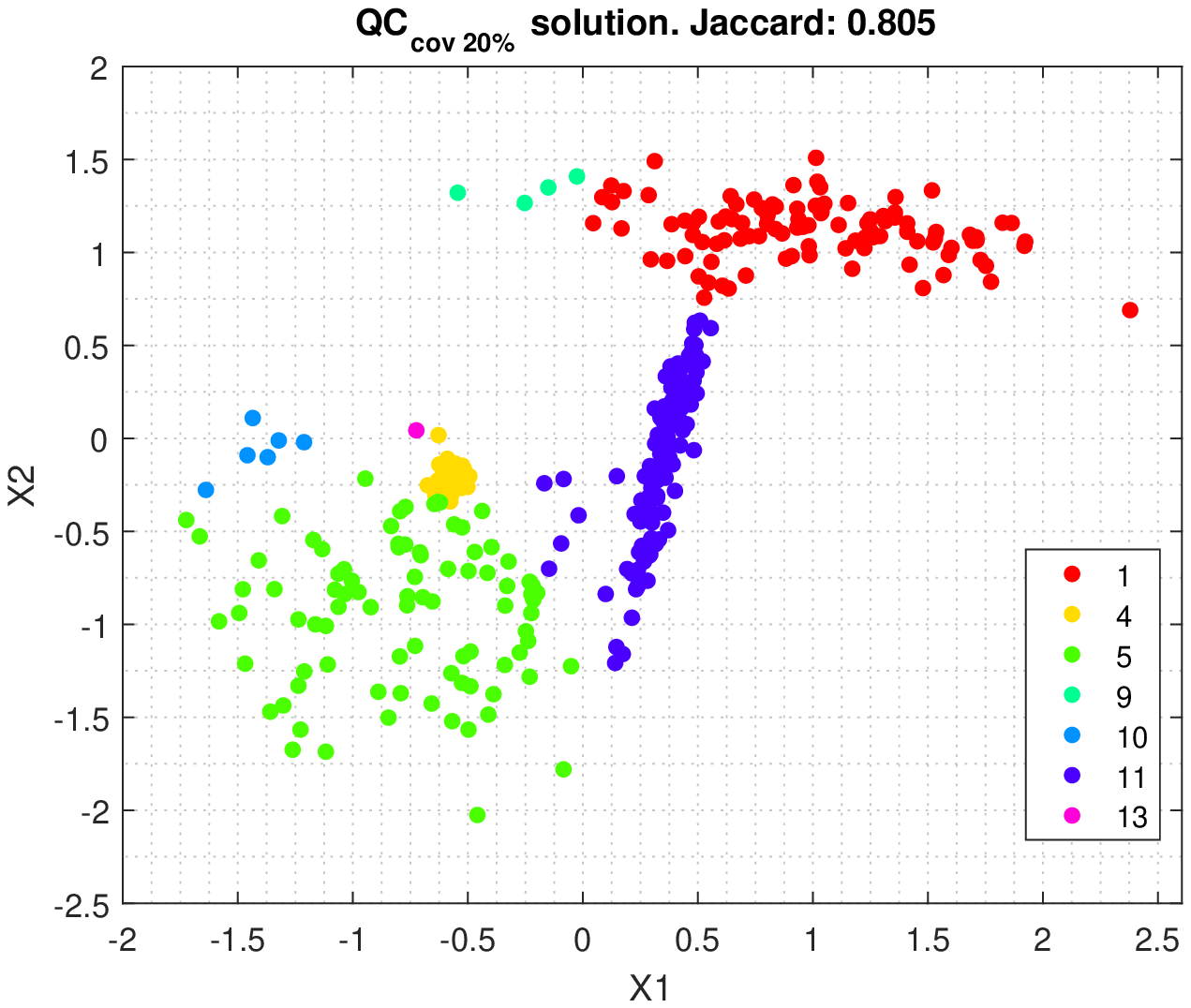}
  \caption{SGD cluster allocation $QC_{cov \, 20\%}$}
  \label{fig:qc3_sol}
\end{subfigure}%
\begin{subfigure}{.5\textwidth}
  \centering
  \includegraphics[width=0.9\linewidth]{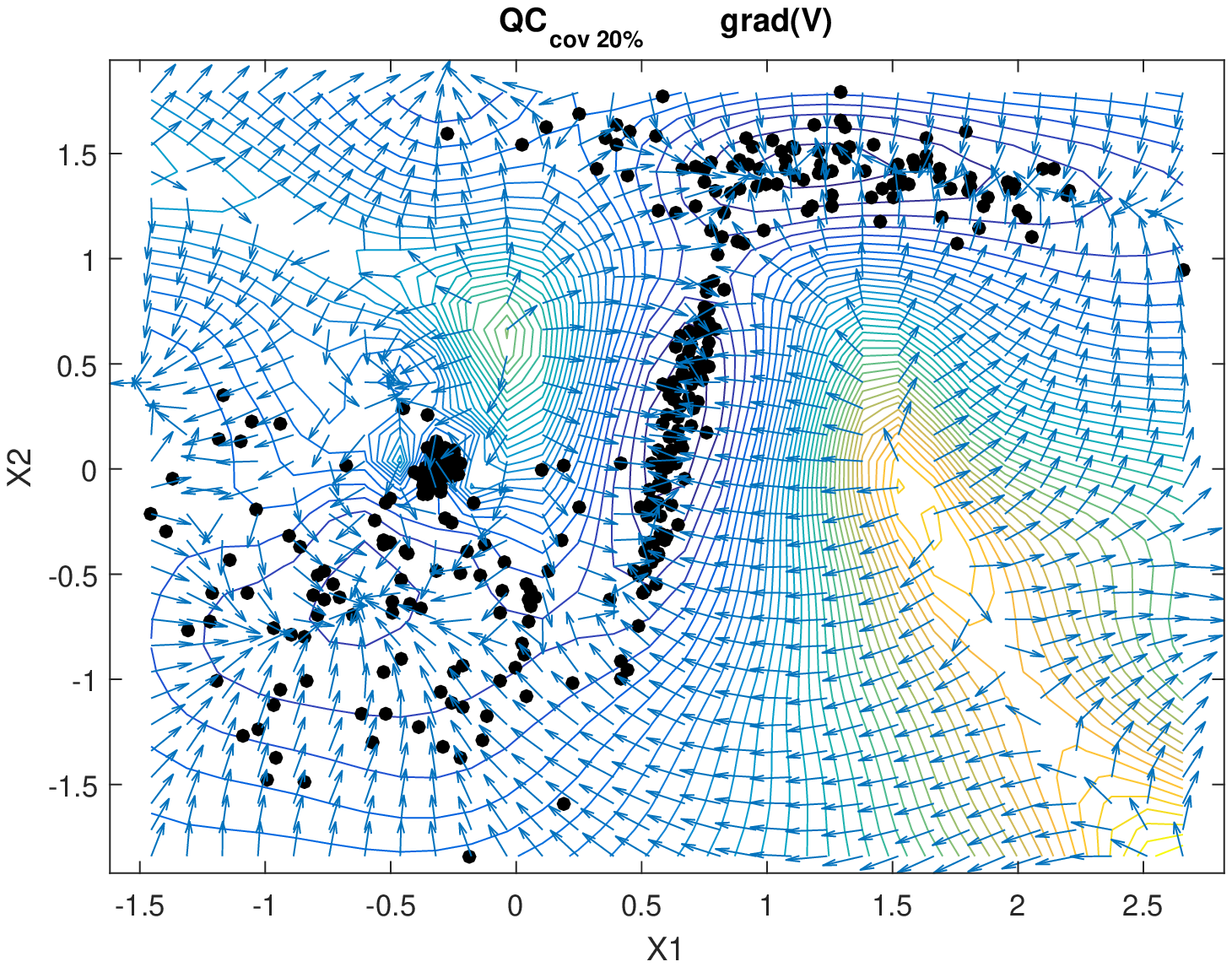}
  \caption{Gradient $QC_{cov \, 20\%}$}
  \label{fig:qc3_grad}
\end{subfigure}
\caption{Cluster allocations by SGD (left) resulting from the gradients of the potential function (right) for artificial data set \#1. The rows correspond to $QC_{\sigma}$, $QC_{knn}$ and $QC_{cov}$ respectively. In all cases the length scales have been computed using a quantile of 20\%. These solutions have Jaccard scores of 0.556, 0.862 and 0.805 respectively.}
\label{fig:qc1_all}
\end{figure*}

Figure~\ref{fig:qc1_sol} shows the cluster allocation by SGD from the potential gradient of the $QC_{\sigma}$ model with a length scale of $\sigma_{20\%}$ showing the corresponding direction of the gradient vectors in figure~\ref{fig:qc1_grad}. The length scale adjusted for all of the data is too broad to accurately capture the high density cluster and too narrow for the sparse cluster at the bottom of the plot which breaks up into multiple local minima. In other words, the wave function, and consequently the potential, are too smooth to fit the local density changes, thus providing a biased clustering. The resulting Jaccard score against the clusters identified by the generating density functions is 0.556. This example is a straightforward example of the difficulties of ${QC}_{\sigma}$ to classify data whose density is locally variable.


\subsection{\texorpdfstring{K-neighbours Quantum Clustering, ${QC}_{knn}$}{}}
\label{QC2}

Information about local density can be included in the length scale by defining $\sigma$ as a function of the KNNs, where the new hyper-parameter is the quantity of neighbours to consider. This quantity will be expressed as a percentage of the total sample size:  $K = \%$KNN.

\begin{eqnarray}
\label{eq:sigmavar}
\sigma_i \equiv \frac{1}{K} \sum_{j \: \in \: knn(\mathbf{x}_i)}^{K} dist(\mathbf{x}_i,\mathbf{x}_j)
\end{eqnarray}

Each observation contributes a different Gaussian function to the overall wave function in equation~(\ref{eq:wave2}). Multiplying through by $\frac{1}{n}$ ensures correct normalisation of the integral over the input space, $\int_{\mathbb{R}^d} \Psi(\mathbf{x})d\mathbf{x} = 1$.

\begin{eqnarray}
\label{eq:wave2}
\Psi(\mathbf{x})= \frac{1}{n} \sum_{i=1}^{n} \psi_i\left(\mathbf{x}\right) =  \frac{1}{n} \sum_{i=1}^{n} \frac{e^{-\frac{\left(\mathbf{x}-\mathbf{x}_i\right)^2}{2\sigma_i^2}}}{\left( \sqrt{2\pi}\sigma_i \right)^{d}}
\end{eqnarray}

\noindent where $d$ is the dimensionality of the sample. One may observe that in~(\ref{eq:shcrodinger_eq}) the total kinetic term is decoupled, $T(\mathbf{x}) = \frac{-\sigma^2}{2}\nabla^2 \Psi(\mathbf{x})$ ($\sigma^2$ and $\nabla^2$ are separated factors because $\sigma$ is a constant common factor), now, with a different $\sigma_i$ per observation, the Schr\"odinger equation~(\ref{eq:shcrodinger_eq}) has to be updated. The kinetic term of each observation, $T_i$, can be expressed as follows:

\begin{equation}\label{eq:kinetic1}
\begin{split}
T_i = \frac{\sigma_i^2}{2} \nabla^2 \psi_i = 
 \left( \frac{\left(\mathbf{x}-\mathbf{x}_i\right)^2}{2\sigma_i^2} - \frac{d}{2} \right) \psi_i
\end{split}
\end{equation}

Therefore, the new total kinetic term couples the length scale $\sigma_i$ and $\nabla^2 \psi_i$:

\begin{equation}\label{eq:kinetic2}
\begin{split}
T_{total} =  \sum_{i=1}^{n} T_i = \sum_{i=1}^{n} \frac{\sigma_i^2}{2} \nabla^2 \psi_i 
\end{split}
\end{equation}

The new potential and its gradient are similar to equations~(\ref{eq:V2}) and~(\ref{eq:gradV2}), but with a variable $\sigma_i$:
\begin{eqnarray}
\label{eq:V3}
V(\mathbf{x})= E + \frac{\sum_{i} \frac{\sigma_i^2}{2} \nabla^2 \psi_i}{\sum_{i} \psi_i} = E - \frac{d}{2} +{\Big \langle \frac{\left(\mathbf{x}-\mathbf{x}_i\right)^2}{2 \sigma_i^2} \Big \rangle}_\Psi
\end{eqnarray}

\begin{equation}\label{eq:gradV3}
\begin{split}
\nabla V(\mathbf{x}) = {\Bigg \langle \frac{\left(\mathbf{x}-\mathbf{x}_i\right)}{\sigma_i^2} \Bigg \rangle}_\Psi \left(1 + \bigg \langle {\frac{\left(\mathbf{x}-\mathbf{x}_i\right)^2}{2\sigma_i^2} \bigg \rangle}_\Psi  \right) - \\
{\Bigg \langle \frac{\left(\mathbf{x}-\mathbf{x}_i\right) \left(\mathbf{x}-\mathbf{x}_i\right)^2}{2 \sigma_i^4}  \Bigg \rangle}_\Psi 
\end{split}
\end{equation}

This approach has been explored in the literature~\cite{ar:nasios2006, inp:nasios2005, casana_qc_nc_2017} to resolve the problem of heteroscedastic data. Using again the artificial data set \#1 as an example, the variable length scale produces a wave function with a very pronounced peak in the high density region causing a ``volcano effect" in the potential (figure~\ref{fig:qc2_grad}). The shape of ${QC}_{knn}$ potential is much more complex than that obtained by ${QC}_{\sigma}$, as it is now smooth in sparse regions and steep in dense areas, as required. Figure~\ref{fig:qc2_sol} shows the cluster allocation by SGD over this potential with accurate discrimination of the high density cluster against the surrounding sparse cluster. The potential also adapts to the local density changes, creating a sharp sink around the highest density peak; this region will be isolated in the clustering allocation by SGD, allowing a cluster discrimination by local densities. In this example, $\sigma_{20\%}$ is an appropriate parameter value; if $\sigma$ were much smaller it would produce an overfitted potential, generating too many sub-clusters. JS in ${QC}_{knn}$ (0.862) is much better than in ${QC}_{\sigma}$ (0.556).

Adjusting the length scale from nearest neighbours is clearly effective for  detecting clusters with very different densities and also to accommodate outliers with smooth and flat gradients that do not lead to an unnecessary fragmentation in low density regions. 
This Section has proposed an approach that partially solves the problem of heteroscedasticy but the amount of neighbours considered in the model is still a hyper-parameter to be determined. There is a trade-off between  too few neighbours resulting in an overfitted density function with too many clusters, and too large a neighbourhood leading to a biased density function with too few clusters.

\subsection{\texorpdfstring{Covariance-based Manifold Quantum Clustering, $QC_{cov}$}{}}
\label{QC3}

Gaussian kernels with non-spherical covariance matrices estimated from local manifold information are proposed in~\cite{vincent2003manifold}. The local covariance matrix, $\Sigma_i$ is computed using the relative distribution of the KNNs around each observation:

\begin{equation}\label{eq:covmat}
\begin{split}
\Sigma_i = \frac{1}{N_k -1} \sum_{j \: \in \: knn}^{N_k} \left(\mathbf{x_j}-\mathbf{x}_i\right)^T \left(\mathbf{x_j}-\mathbf{x}_i\right)
\end{split}
\end{equation}

Now, each observation has a kernel with the form of a multivariate normal distribution, producing the following wave function:

\begin{equation}
\label{eq:wavecov}
\begin{split}
\Psi(\mathbf{x})= \frac{1}{n} \sum_{i=1}^{n} \psi_i\left(\mathbf{x}\right) = \\
 \frac{1}{n} \sum_{i=1}^{n} \frac{1}{\sqrt{|2\pi\Sigma_i|}} e^{-\frac{1}{2}\left(\mathbf{x}-\mathbf{x}_i\right)^T \Sigma_i^{-1} \left(\mathbf{x}-\mathbf{x}_i\right)}
\end{split}
\end{equation}

This wave function is a more accurate probability density function than those presented in Sections \ref{origQC} and \ref{QC2} since each observation captures the distribution of the nearest neighbours. The density function reproduces faithfully elongated distributions, like cigar shapes or even spiral shapes, a case studied in Section~\ref{Res}. But it also has some disadvantages:

\begin{enumerate}
\item The mathematical complexity of the potential function and the corresponding gradients is significantly increased.
\item Degenerate covariance matrices i.e. with diagonal elements close to zero, may cause singularities in the covariance-inverse estimation.
\item If the covariances are too anisotropic, the positive effect of superposition in the wave function is considerably reduced. This produces a wave function that is less smooth and a potential less convex, favouring the creation of an excessive number of local minima if the Gaussian kernels do not overlap one another enough.
\end{enumerate}

These disadvantages can be mitigated if all the local covariance matrices are restricted to be diagonal even if anisotropic (expressed into their eigenvector basis). Degeneracy is avoided by setting a minimum threshold value for the diagonal elements, as shown in Appendix~\ref{app:selcovthreshold}. This also improves the superposition effect, because all Gaussian kernels have a minimum radius controlled by the local-covariance threshold, but a larger ellipsoid axis greater when needed. The local-covariance threshold is defined by:

\begin{eqnarray}
\label{eq:covthreshold}
\sigma_{th_i}^2 = \frac{\sigma_{k'nn_i}^2}{d}
\end{eqnarray}

\noindent where $d$ is the dimension of the data and $\sigma_{k'nn_i}$ the mean distance of the k' NNs of each observation $i$, namely, the variable length scale used in $QC_{k'nn}$. The percentage of neighbours considered, $k'$, is determined experimentally in appendix~\ref{app:selcovthreshold}; results show that $k'$ should be the same $k$ used to compute the local covariance matrix, in order to keep enough interaction between kernels; this means that the solution may well be a hybrid model between $QC_{knn}$ and $QC_{cov}$.\par

The Schr\"odinger equation~(\ref{eq:shcrodinger_eq}) must be adapted to the variable length scale $\sigma_i$, again. In this case, $\sigma_i$ must be replaced by a scalar expression of $\Sigma_i$; considering that $\Sigma_i$ is diagonal if it is expressed in an eigenvector basis, it is possible to make the change $\sigma_i^2  \to  tr(\Sigma_i)$:\par

\begin{equation}\label{eq:Vcov}
\begin{split}
V(\mathbf{x}) = E + \Big \langle  \frac{tr(\Sigma_i)}{2}  tr\left( \Sigma_i^{-1} \left(\mathbf{x}-\mathbf{x}_i\right)  \left(\mathbf{x}-\mathbf{x}_i\right)^T  \Sigma_i^{-1} \right)  \Big \rangle_\Psi \\
 - \Big \langle  \frac{1}{2}tr\left(\Sigma_i \right)  tr\left(\Sigma_i^{-1} \right)  \Big \rangle_\Psi
\end{split}
\end{equation}

The sample local covariance estimate is diagonalized to threshold eigenvalues to a  minimum value that  is higher than a small regularization term, making further regularization
unnecessary. This imposes a minimum radius in the covariance matrix resulting in $QC_{cov}$ that are similar in value to $QC_{knn}$. In contrast, the gradient and equipotential surfaces of figure~\ref{fig:qc3_grad} show contours that better capture anisotropy. However, the $QC_{cov}$ potential is less smooth than $QC_{knn}$ potential and tends to create more local minima that produce sub-clusters as seen in figure~\ref{fig:qc3_sol}, which shows the allocation of seven clusters. The occurrence of these sub-clusters reduces performance (JS=0.805) motivating the proposal made in Section~\ref{PQC}.

\subsection{\texorpdfstring{Probabilistic Quantum Clustering, ${QC}_{cov}^{prob}$}{}}
\label{PQC}

PQC, ${QC}_{cov}^{prob}$ modifies the cluster allocation of ${QC}_{cov}$; in particular, the clusters are no longer defined by the groups of points found after the SGD. These groups are now used to define component elements (subfunctions) that add to make the overall wave function.

The starting point for the probabilistic framework ${QC}_{cov}^{prob}$ is to attribute the joint probability of observing cluster $k$ in the position $\mathbf{x}$ to the sum of Gaussian functions associated with the observations grouped in the cluster (subfunction) $k$:

\begin{eqnarray}
\label{eq:probx}
\Psi(\mathbf{x})= \sum_{k=1}^{K} \frac{\sum_{i \in k}^{\#k} \psi_i\left(\mathbf{x}\right)}{n} = \sum_{k=1}^{K} P(k,\mathbf{x}) = P(\mathbf{x})
\end{eqnarray}

\noindent where $n$ is the sample size, $K$ the total number of clusters, and $\#k$ the number of observations in cluster $k$.

Equation~(\ref{eq:probx}) could be seen as a generative model using a mixture of Gaussians, one centred at each observation, with prior probability equal to $1/n$ (isotropic Gaussian kernel over the data). The purpose of the QC is to provide a link between the individual Gaussian functions so that points in the same cluster are linked together.

The probability of $k$ can be obtained by marginalizing the joint probability over $\mathbb{R}$:

\begin{equation}
\label{eq:probk}
\begin{split}
P(k) = \int_{\mathbb{R}} P(k,\mathbf{x}) d\mathbf{x} = \int_{\mathbb{R}} \frac{\sum_{i \in k}^{\#k} \psi_i\left(\mathbf{x}\right)}{n} d\mathbf{x} = \\
 \sum_{i \in k}^{\#k} \frac{\int_{\mathbb{R}} \psi_i\left(\mathbf{x}\right)d\mathbf{x} }{n} = \sum_{i \in k}^{\#k} \frac{1}{n} = \frac{\#k}{n}
\end{split}
\end{equation}

Once the joint probability is defined, the Bayes' rule can be applied to obtain the following probabilities:

\begin{eqnarray}
\label{eq:probk_x}
P(k|\mathbf{x}) = \frac{P(k,\mathbf{x})}{P(\mathbf{x})} = \frac{\sum_{i \in k}^{\#k} \psi_i\left(\mathbf{x}\right)}{\sum_{k=1}^{K} \sum_{i \in k}^{\#k} \psi_i\left(\mathbf{x}\right)}
\end{eqnarray}

\begin{eqnarray}
\label{eq:probx_k}
P(\mathbf{x}|k) = \frac{P(k,\mathbf{x})}{P(k)} = \frac{\sum_{i \in k}^{\#k} \psi_i\left(\mathbf{x}\right)}{\frac{\#k}{n}}
\end{eqnarray}

Now, $P(k|\mathbf{x})$ can be used to define a new probabilistic cluster allocation:

\begin{eqnarray}
\label{eq:cl_alloc}
cluster(\mathbf{x}) = \operatorname{arg\,max}_k P(k|\mathbf{x})
\end{eqnarray}

In other words, the cluster allocation in the region $\mathbf{x}$ will correspond to the cluster $k$ such as $\operatorname{arg\,max}_k P(k|\mathbf{x})$, or equivalently $\operatorname{arg\,max}_k P(k,\mathbf{x})$ since $P(\mathbf{x})$ is a common denominator.\par

Summing up, there are two cluster allocations in the algorithm pipeline:\par

The first one is when the gradient descent is performed over the potential to allocate each observation of the training set into its corresponding potential well (identified as cluster). This gives the grouped Gaussians per cluster $k$; according to equation~(\ref{eq:probk}), the probabilistic framework can be derived: $P(k,\mathbf{x}) = \frac{1}{n} \sum_{i \in k}^{\#k} \psi_i\left(\mathbf{x}\right)$

The second cluster allocation, called probabilistic cluster allocation, is based on the probabilistic framework, and it only decides to which cluster each observation belongs based on the $P(k|\mathbf{x})$, selecting k so that makes $P(k|\mathbf{x})$ maximum. 
As a consequence of this, it is possible that the model generates $k$ clusters but not all of them contain observations. In other words, there would exist $k'$ empty (small) clusters if $P(k'|\mathbf{x})$ never wins in any region of the input space, $\mathbf{x}$.

This is a significant improvement over the original method for cluster allocation because any region of the input space can be allocated to a cluster without the need to apply SGD over the potential. The probabilistic cluster allocation draws a probability map based on $P(k|\mathbf{x})$ to define the boundaries between clusters. Although the probability map still requires one cluster allocation by SGD, no additional SGD is required for new observations.

Experimental results show a difference lower than 2\% in the cluster allocation when comparing ${QC}_{knn}$ with its probabilistic counterpart. Differences are greater in the case of $QC_{cov}$, with the probabilistic cluster allocation closer to the true labels than the SGD approach, with JS=0.882, i.e., the highest JS among all the experiments carried out, and with the selection of only four of the seven clusters detected in figure~\ref{fig:qc3_sol}.

Another interesting characteristic of the probabilistic approach is the capability to implement outlier detection where the probability of belonging to any cluster is lower than a given threshold ($P(\mathbf{x}|k) < threshold$).
Therefore, $P(\mathbf{x}|k)$ and $P(k|\mathbf{x})$ map the probability functions of belonging to each cluster and the regions formed by outliers. 

\begin{figure*}
\centering
\begin{subfigure}{.5\textwidth}
  \centering
  \includegraphics[width=0.9\linewidth]{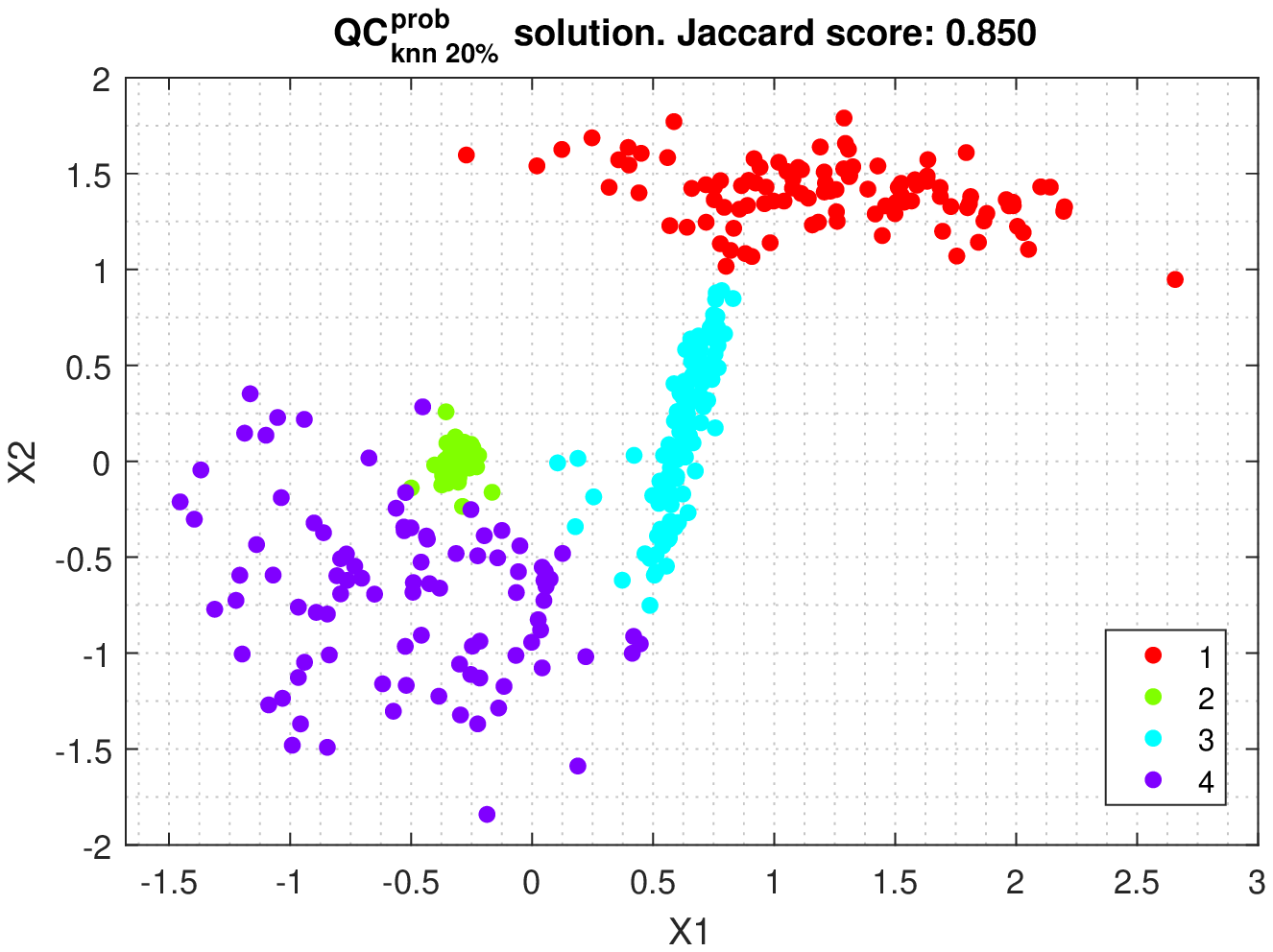}
  \caption{Solution for $QC^{prob}_{knn \, 20\%}$}
  \label{fig:data1_qc2_sol_prob}
\end{subfigure}%
\begin{subfigure}{.5\textwidth}
  \centering
  \includegraphics[width=0.9\linewidth]{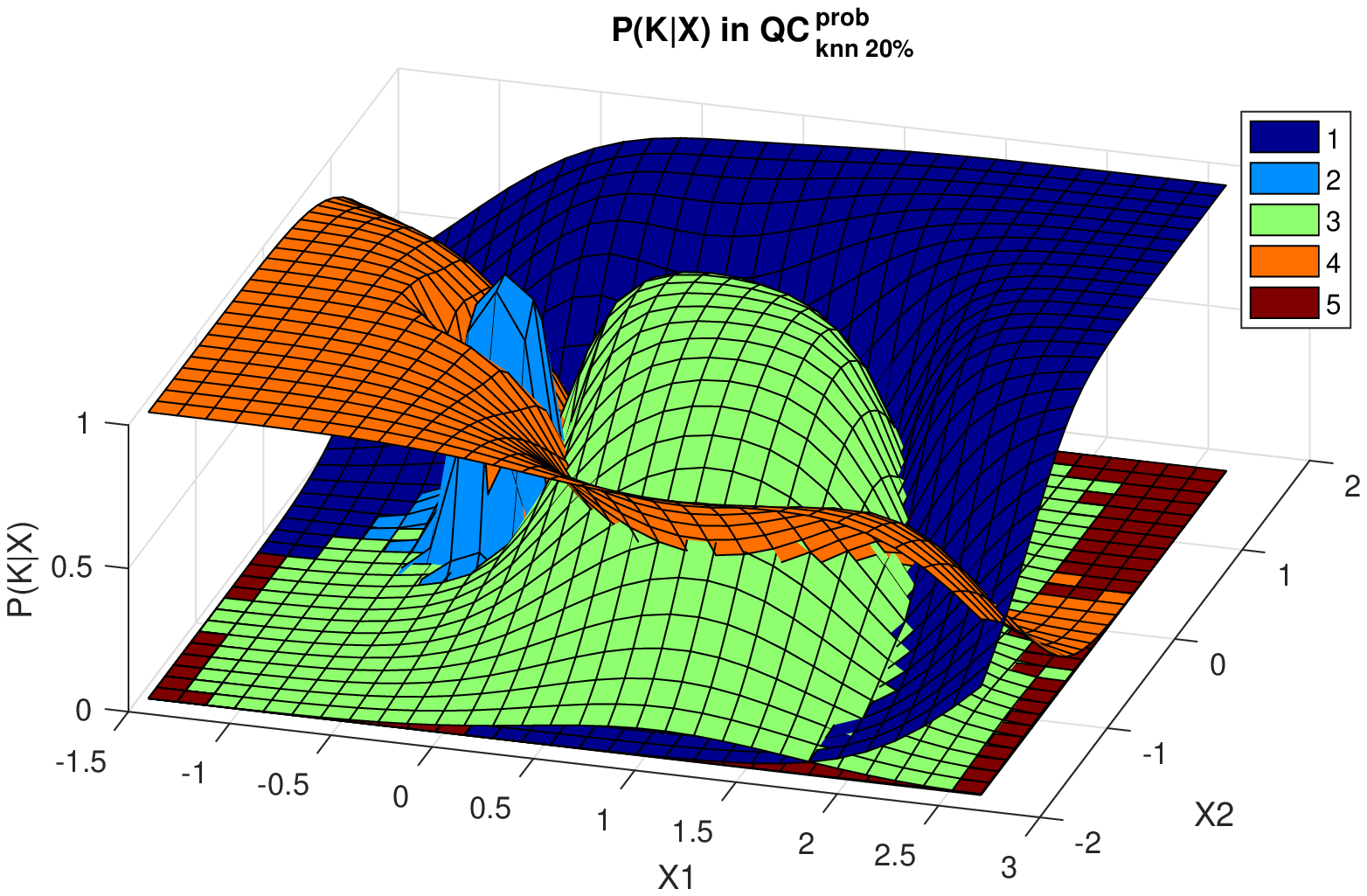}
  \caption{$P(K|X)$}
  \label{fig:data1_qc2_Pkx_20}
\end{subfigure}
\begin{subfigure}{.5\textwidth}
  \centering
  \includegraphics[width=0.9\linewidth]{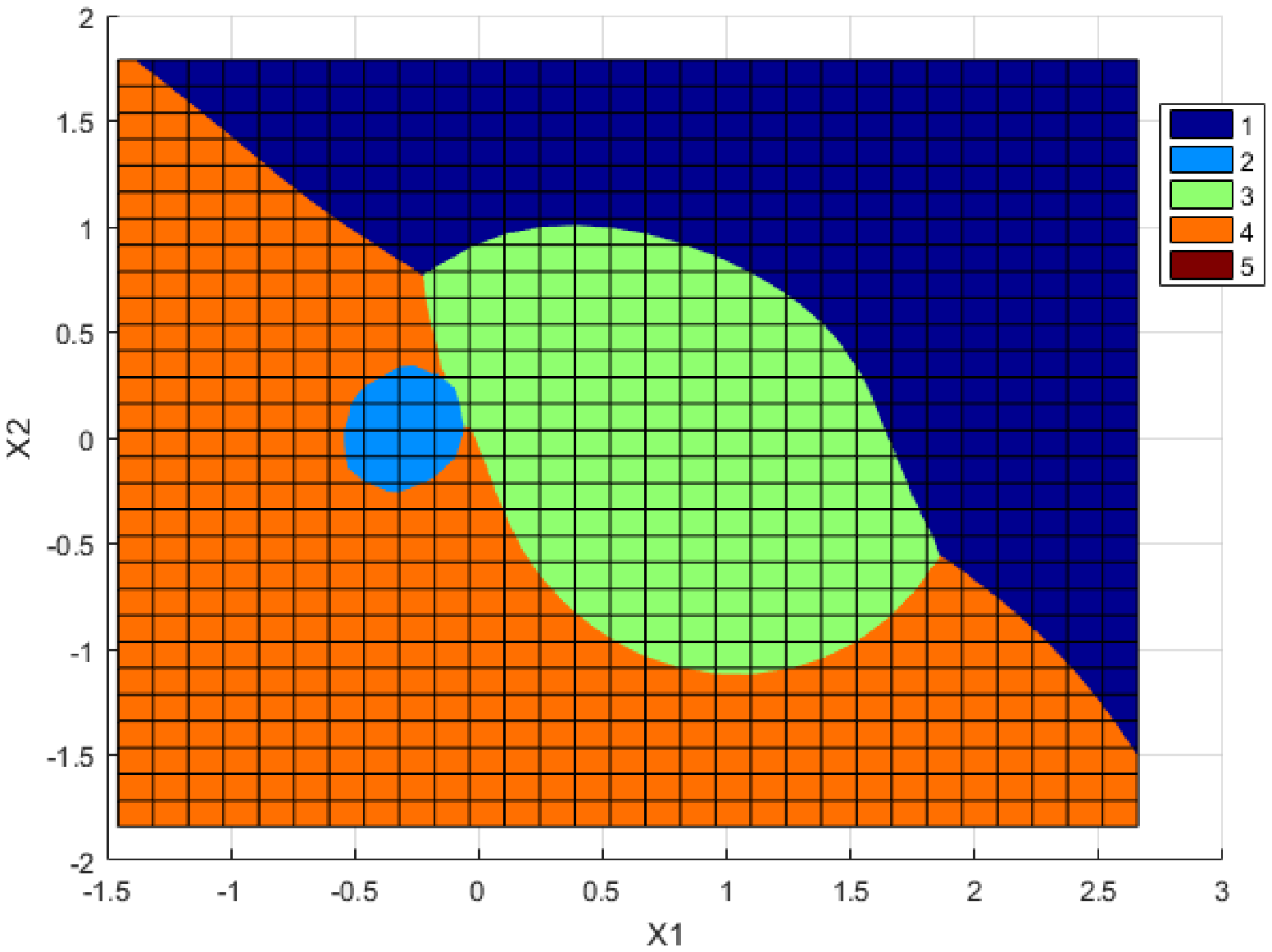}
  \caption{Projection of $P(K|X)$}
  \label{fig:data1_qc2_Pkx_20_2D_screen}
\end{subfigure}%
\begin{subfigure}{.5\textwidth}
  \centering
  \includegraphics[width=0.9\linewidth]{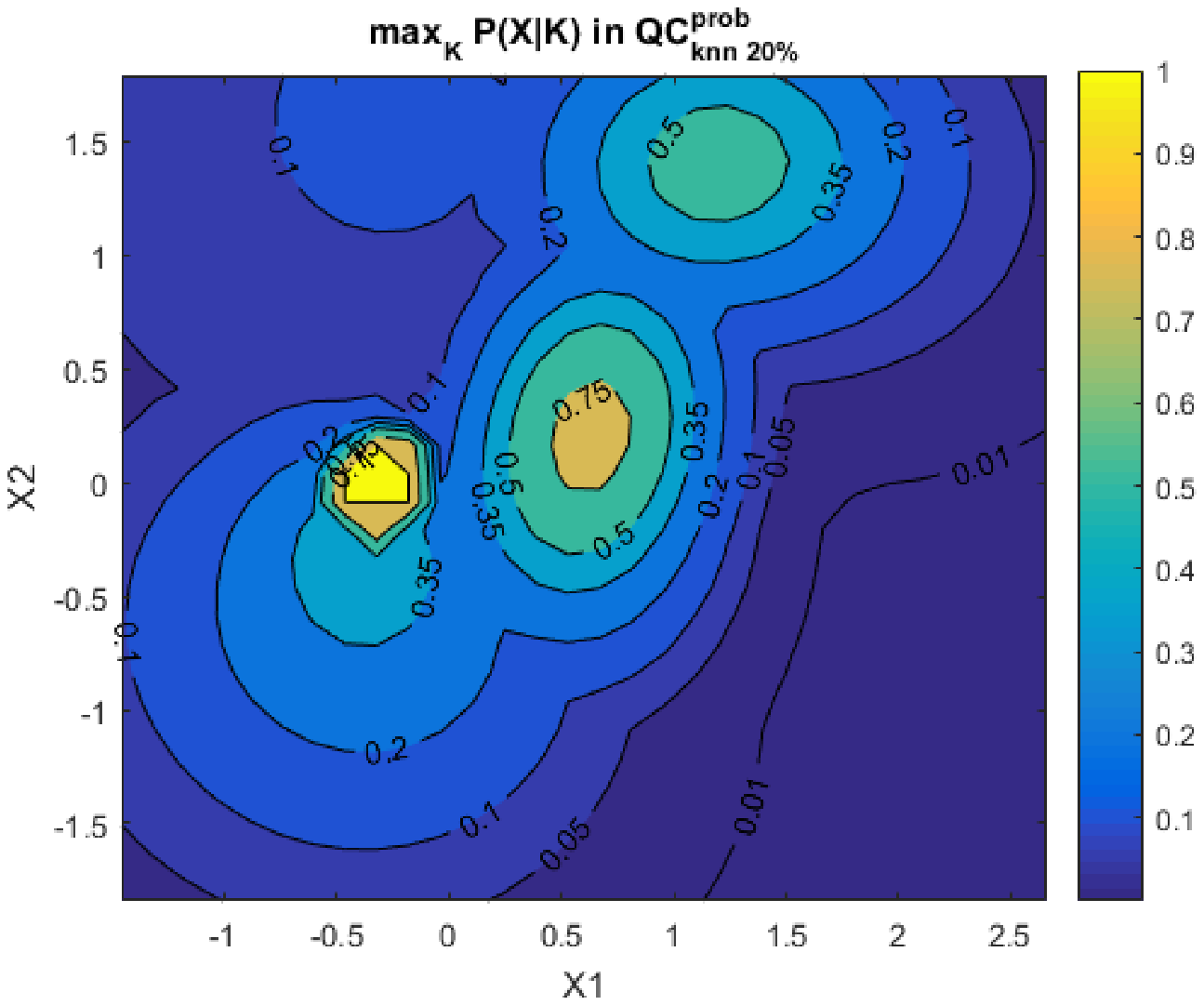}
  \caption{$max_K \, \, P(X|K)$}
  \label{fig:data1_qc2_Pxk_20_heatmap}
\end{subfigure}
\caption{Top left figure shows the probabilistic cluster allocation with $QC^{prob}_{knn \, 20\%}$ (JS=0.850). Top right figure shows its probability map of cluster membership, $P(K|X)$. A top-down projection can be observed in bottom left figure, where only the highest cluster membership regions are observed. The bottom right figure depicts $max_K \, \, P(X|K)$, which is useful for outlier detection.}
\label{fig:data1_prob}
\end{figure*}

Figures~\ref{fig:data1_qc2_Pkx_20} and~\ref{fig:data1_qc2_Pkx_20_2D_screen} depict the probability maps using the five clusters detected in the ${QC}_{knn}$ solution (figure~\ref{fig:data1_qc2_sol_prob}); there is a small cluster (brown colour) that is covered by other clusters with higher probabilities. Figure~\ref{fig:data1_qc2_Pkx_20_2D_screen} (zenital view of the probability map) shows that clusters can be allocated without using SGD.
Figure~\ref{fig:data1_qc2_Pxk_20_heatmap} shows the maximum probability, $P(X|K)$, the model can assign to each region, becoming a tool for outlier detection.

Something similar happens with the solutions of ${QC}_{cov}^{prob}$, where the algorithm detects seven clusters, but the probabilistic allocation makes a reduction to four clusters.

\subsection{Performance assessment}
\label{ANLL}
\subsubsection{Average Negative Log-Likelihood, ANLL}

This section deals with a major contribution of the paper, namely, an unsupervised method to select the remaining free hyper-parameter: \%KNN. As each observation is allocated to the cluster $k$ with the highest $P(k|\mathbf{x})$, cluster $k_w$, the $i$-th observation is allocated with a probability $P(k_w|\mathbf{x}_i)$.\par
$P(k_w|\mathbf{x}_i)$ competes against the probabilities associated with the other clusters in $\mathbf{x}_i$. Rewriting equation~(\ref{eq:probk_x}) in terms of probability marginalization:

\begin{eqnarray}
\label{eq:probk_x2}
P(k_w|\mathbf{x}) = \frac{P(k_w,\mathbf{x})}{P(\mathbf{x})} = \frac{P(k_w,\mathbf{x})}{\sum_k P(k,\mathbf{x})}
\end{eqnarray}

Regions close to the cluster boundaries have low values of $P(k_w,\mathbf{x})$. Therefore, the best models have the highest value of $P(k_w|\mathbf{x})$ for a high number of observations. This corresponds to the likelihood of cluster membership, given by:

\begin{eqnarray}
\label{eq:LL}
LL(K|\mathbf{X}) = log \left( \prod_i^n P(k_w|\mathbf{x}_i) \right) = \sum_i^n log\left( P(k_w|\mathbf{x}_i) \right)
\end{eqnarray}

To normalize the score in the range $[0,1]$, the average negative log-likelihood (ANLL) is used:

\begin{eqnarray}
\label{eq:ANLL}
ANLL(K|\mathbf{X}) = \frac{-\sum_i^n log\left( P(k_w|\mathbf{x}_i) \right)}{N} 
\end{eqnarray}

The lower the ANLL, the better the model fit. Its value clearly depends on the length scale parameter, \%KNN, because the length scale controls the number of clusters and the smoothness of the wave function. A representation of ANLL against \%KNN will, in general, have some regions where the ANLL score is minimized, so that $P(k_w|\mathbf{x})$ is maximized, obviously avoiding the trivial solution of a single cluster covering all of the data, which takes the lowest possible value (ANLL = 0).

\begin{figure}[!ht]
\centering
\includegraphics[width=1\linewidth]{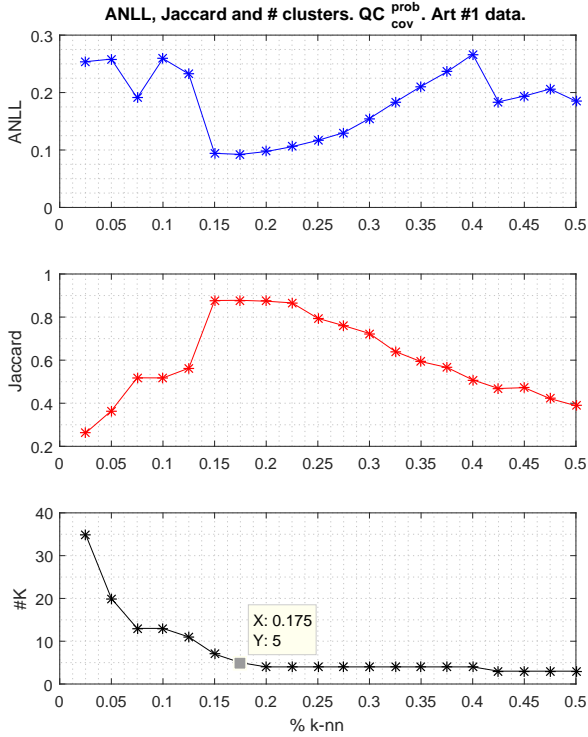}
\caption{Comparative plot of ANLL against Jaccard score as function of $\%$KNN, using ${QC}_{cov}^{prob}$ in artificial data set \#1. The lower values of ANLL coincide with the higher values of Jaccard score. Also ANLL points out how the structure of the data is changing when \%KNN varies. The bottom figure depicts the number of clusters per length scale solution.}
\label{fig:data1_qc3_knn_js_k}
\end{figure}

The ANLL provides an unsupervised figure of merit which is highly correlated with the supervised JS. Therefore, it can be used as a measure of the clustering performance without the need of prior information about the number of clusters or their composition. Figure~\ref{fig:data1_qc3_knn_js_k} shows ANLL and JS for different length scales in $QC_{cov}^{prob}$, to illustrate their correlation. In addition, the ANLL vs \%KNN plot reveals the hierarchical structure of the data, where an abrupt change in ANLL means a significant change in the data structure. The bottom plot of figure~\ref{fig:data1_qc3_knn_js_k} shows how the number of clusters depends on the length scale, although the $QC_{cov}^{prob}$ considerably cushions the fluctuation compared with the original QC.

\medskip

For the artificial data set \#1 and $QC_{cov}^{prob}$, the Pearson's linear correlation coefficient between Jaccard score and ANLL is $\rho = -0.776$, p-value $<0.001$.

\subsubsection{Extended ANLL score}
\label{sec:extended_ANLL}
The extended ANLL score improves ANLL by setting a threshold $E_{th}$ to merge two clusters according to the maximum potential difference between their centroids. A more detailed explanation is presented in appendix~\ref{app:c_alloc}.

By default, the ANLL score  uses a fixed $E_{th}$ that depends on the SGD convergence criteria in the last iteration:
\begin{eqnarray}
\label{eq:Eth_default}
E_{th} =  max\Big( \epsilon_V, max\big(\Delta V(\mathbf{x}_{iter \, max})\big)\Big)
\end{eqnarray}
\noindent This $E_{th}$ takes the maximum value between the minimum SGD precision, $\epsilon_V$, and the last SGD update, in such a way it corresponds with the lowest possible bound, so that any accidental cluster merging is avoided.

\begin{figure}[!ht]
\centering
\includegraphics[width=1\linewidth]{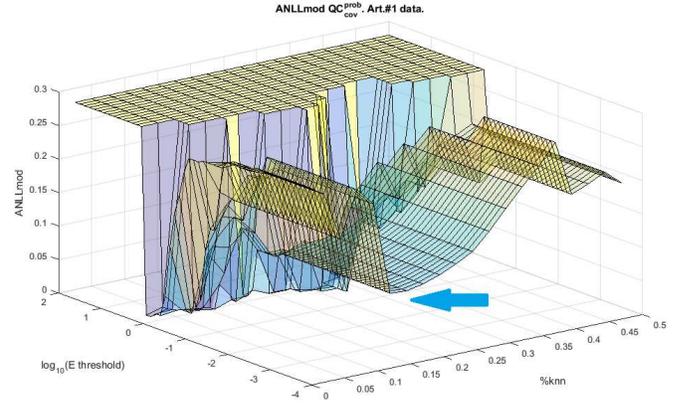}
\caption{Extended ANLL score versus $\%$KNN and $E_{th}$ for data set \#1. The ANLL scores always diminish when $E_{th}$ increases, because it is an implicit reduction of the number of clusters. That explains why the ANLL score is less reliable when there are few clusters, because the trivial solution, with a unique cluster, always leads to ANLL equal to zero. To avoid confusion with non-trivial solutions, scores associated with a trivial solution are assigned with the highest ANLL score.}
\label{fig:data1_qc3_ANLLmod}
\end{figure}

Figure~\ref{fig:data1_qc3_ANLLmod} shows an enhanced representation of ANLL, including its relationship with $E_{th}$. To avoid confusion with non-trivial solutions, scores associated with a trivial solution are assigned to the highest ANLL score.

The interpretation of the ANLL plots is partly subjective, as the plots give an indication of the clustering structure in the data which may be multi-level when the data are hierarchical. The following steps must be taken:

\begin{enumerate}
\item Look for a local minimum in the direction of $\%$KNN axis giving priority to the lowest values of $\%$KNN.

\item Local minima must have a stable valley in the direction of $E_{th}$, any solution within this valley is a good solution.

\item Repeat the process if there are more local minima in the direction of $\%$KNN, in ascending order.

\item Looking at the $E_{th}$ direction, if there is a stable region of low ANLL values, wide enough to cover several values of $E_{th}$ and $\%$KNN, that region contains a meaningful solution. Solutions with a high $E_{th}$ must be taken with caution because they might correspond to solutions with a few number of clusters, produced after merging clusters hierarchically.

\end{enumerate}

Figure~\ref{fig:data1_qc3_ANLLmod} shows extended ANLL score versus $E_{th}$ and $\%$KNN. For low $E_{th}$ values, being $E_{th} = 0.001$ the default value, the figure~\ref{fig:data1_qc3_ANLLmod} presents the same pattern of ANLL observed in figure~\ref{fig:data1_qc3_knn_js_k}. 

\section{Data sets}
\label{data}

Two challenging artificial data sets and two real-world data sets were employed to test the theoretical hypotheses and evaluate the clustering performance.

\subsection{Data set \#1 (artificial): Local densities}
\label{art1}
This data set has two main characteristics which challenge clustering algorithms: first, there are two clusters with cigar shapes; second, there are two clusters partially overlapped but with different local densities. The original QC was able to detect anisotropic clusters, but it is less able to discriminate clusters with different local densities. 
The data set is two-dimensional to aid visualization and comprises four clusters with 100 observations each.

\subsection{Data set \#2 (artificial): Two spirals}
\label{spirals}

This is a two-dimensional spiral data set with standard deviation in the first spiral of 0.1 and 0.025 in the second spiral. Each cluster has 200 observations.

\subsection{Data set \#3 (real): Crabs}
\label{crabs}
This well-known data set was used in the original QC paper,~\cite{NIPS2001_2083}. The Crabs' data set describes five morphological measurements on 50 crabs of each of two colour forms and both sexes, of the species Leptograpsus variegatus collected at Fremantle, W. Australia. In total there are 200 observations and four different labels, two for gender and two for each species. To compare the results with the original paper, principal component analysis (PCA) has been applied, selecting only the two first principal components (PCs).

\subsection{Data set \#4 (real): Olive oil}
\label{olive}

The \textit{Italian olive oil} data set~\cite{forina83} consists of 572 observations and 10 variables. Eight variables describe the percentage composition of fatty acids found in the lipid fraction of these oils, which is used to determine their authenticity. The remaining two variables contain information about the classes, which are of two kinds: three ``super-classes'' at country level: North, South, and the island of Sardinia; \textit{and} nine collection area classes: three from the Northern region (Umbria, East and West Liguria), four from the South (North and South Apulia, Calabria, and Sicily), and two from the island of Sardinia (inland and coastal Sardinia). The hierarchical structure of this data set makes it especially appealing for testing clustering algorithms.

%

\section{Results}
\label{Res}

This section evaluates the extent to which the ANLL score can determine the most suitable $\%$KNN to maximize the JS, highlighting the peculiarities of each data set and comparing the results of both models, $QC_{knn}^{prob}$ and $QC_{cov}^{prob}$. As ANLL tends to be smaller as the number of clusters decreases, when several local minima appear in ANLL, the ones associated with lower $\%$KNN values should have priority over the ones with higher $\%$KNN values.

The tables of results include the following information:

\begin{itemize}
\item Column 1: data set number and QC model.

\item Column 2: score employed to select the quantile (\%KNN), firstly the supervised choice according to the best JS, then the unsupervised option based on the local minima found in ANLL, and finally checking if the extended ANLL has a stable region increasing the $E_{th}$ parameter.

\item Column 3: the $E_{th}$ parameter; by default is used $E_{th}=0.001$, but then the extended ANLL plot is analysed to find stable ANLL regions with solutions of higher hierarchical order.

\item Column 4: length scale parameter in quantiles (\%KNN)

\item Column 5: number of clusters (\#K)

\item Column 6: ANLL score

\item Column 7: Jaccard score - for the Olive oil data there are two possible classifications, with 3 regions or 9 subregions of Italy.

\item Column 8: Cramer\'s V score - for the Olive oil data there are two possible classifications, as above.

\item Column 9: Pearson's linear correlation coefficient between ANLL score with $E_{th}=0.001$ and the Jaccard score. 

\item Column 10: The p-values of correlation coefficient for testing the null hypothesis of no correlation against the alternative that there is a non-zero correlation. 
\end{itemize}

\subsection{Data set \#1: Local densities}
\label{R_art1}

Table~\ref{table1} shows that both models, $QC_{knn}^{prob}$ and $QC_{cov}^{prob}$, perform similarly for this data set. $QC_{knn}^{prob}$ has the correct number of clusters, four, with a $JS = 0.85$, however $QC_{cov}^{prob}$ with five clusters has a slightly better value, $JS = 0.88$. In both cases, the ANLL corresponds with the Jaccard score. On the other hand, there is not a stable region of low ANLL with high $E_{th}$ values, so no hierarchical solution was considered. As complementary information, applying TDA methods with a wide range of parameters, their best result has  lower performance than $QC^{prob}$, with a $JS = 0.60$.

\begin{table*}[!t]
\renewcommand{\arraystretch}{1.3}
\caption{Data set \#1: Local densities. The supervised solution with best JS matches the unsupervised solution proposed by ANLL.}
\label{table1}
\centering
\begin{tabular}{|c|c|c|c|c|c|c|c|c|c|}
\hline
Data \#1                      & \textit{Score}                 & $E_{th}$   & \textit{\%KNN} & \textit{\#K} & \textit{ANLL}   & \textit{JS}    & $C_v$ & $\rho_{E_{th}}$ & \textit{p-val} \\ \hline
\multirow{3}{*}{$QC_{knn}^{prob}$} & Best JS                   & 0.001 & 17.5 & 4  & 0.082 & 0.85 & 0.94  & -0.81           & 1.5E-5   \\ \cline{2-10} 
                                 & Best ANLL                 & 0.001 & 17.5 & 4  & 0.082 & 0.85 & 0.94  & -                & -          \\ \cline{2-10} 
                                 & ANLL stable high $E_{th}$ & No    & -    & -  & -      & -     & -      & -                & -          \\ \hline
\multirow{3}{*}{$QC_{cov}^{prob}$} & Best JS                   & 0.001 & 17.5 & 5  & 0.092 & 0.88 & 0.96  & -0.87           & 6.0E-7   \\ \cline{2-10} 
                                 & Best ANLL                 & 0.001 & 17.5 & 5  & 0.092 & 0.88 & 0.96  & -                & -          \\ \cline{2-10} 
                                 & ANLL stable high $E_{th}$ & No    & -    & -  & -      & -     & -      & -                & -          \\ \hline
\end{tabular}
\end{table*}

\subsection{Data set \#2: Two spirals}
\label{R_spirals}

\begin{figure}[!ht]
\centering
\includegraphics[width=1\linewidth]{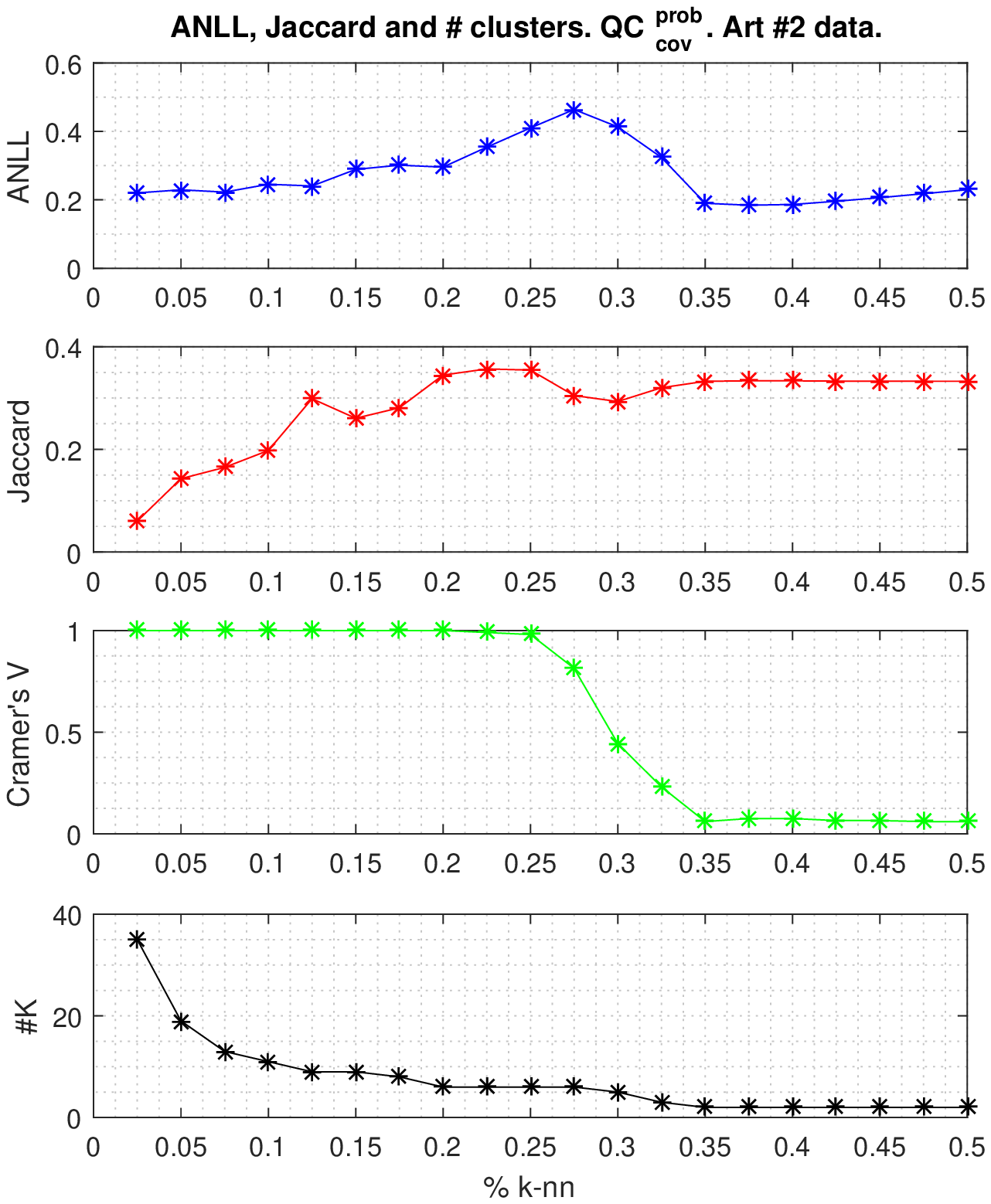}
\caption{ANLL, Jaccard score, $C_v$ and number of clusters obtained by $QC_{knn}^{prob}$ for data set \#2. ANLL splits the graph into two regions separated by a value of KNN equal to $22.5\%$; at the left side, the spirals are not mixed but broken up; while at the right side the spirals are mixed but there are only two clusters. Obviously, an external supervision would prefer not-mixed spirals.}
\label{fig:data2_qc3_knn_js_cv_k}
\end{figure}

Figure~\ref{fig:data2_qc3_knn_js_cv_k} shows that JS is quite low. Actually, JS is not a good metric for this data set as it does not attribute any importance to the fact that the spirals are not mixed, it only measures similarity with the true labels. To address this issue, the Cramer's V-index ($C_v$) was used, which is a normalized version of the standard chi-square test for contingency tables; $C_v$ measures the concordance between different cluster allocations, detecting when the spirals are mixed if $C_v<1$.

$C_v$ shows that the spirals are not mixed until $25\%$ KNN for $QC_{cov}^{prob}$, but they are fragmented into sub-clusters. Length scales greater than $25\%$ KNN make the potential too smooth and the potential wells mix the spirals.

\begin{figure*}
\centering
\begin{subfigure}{.5\textwidth}
  \centering
  \includegraphics[width=0.9\linewidth]{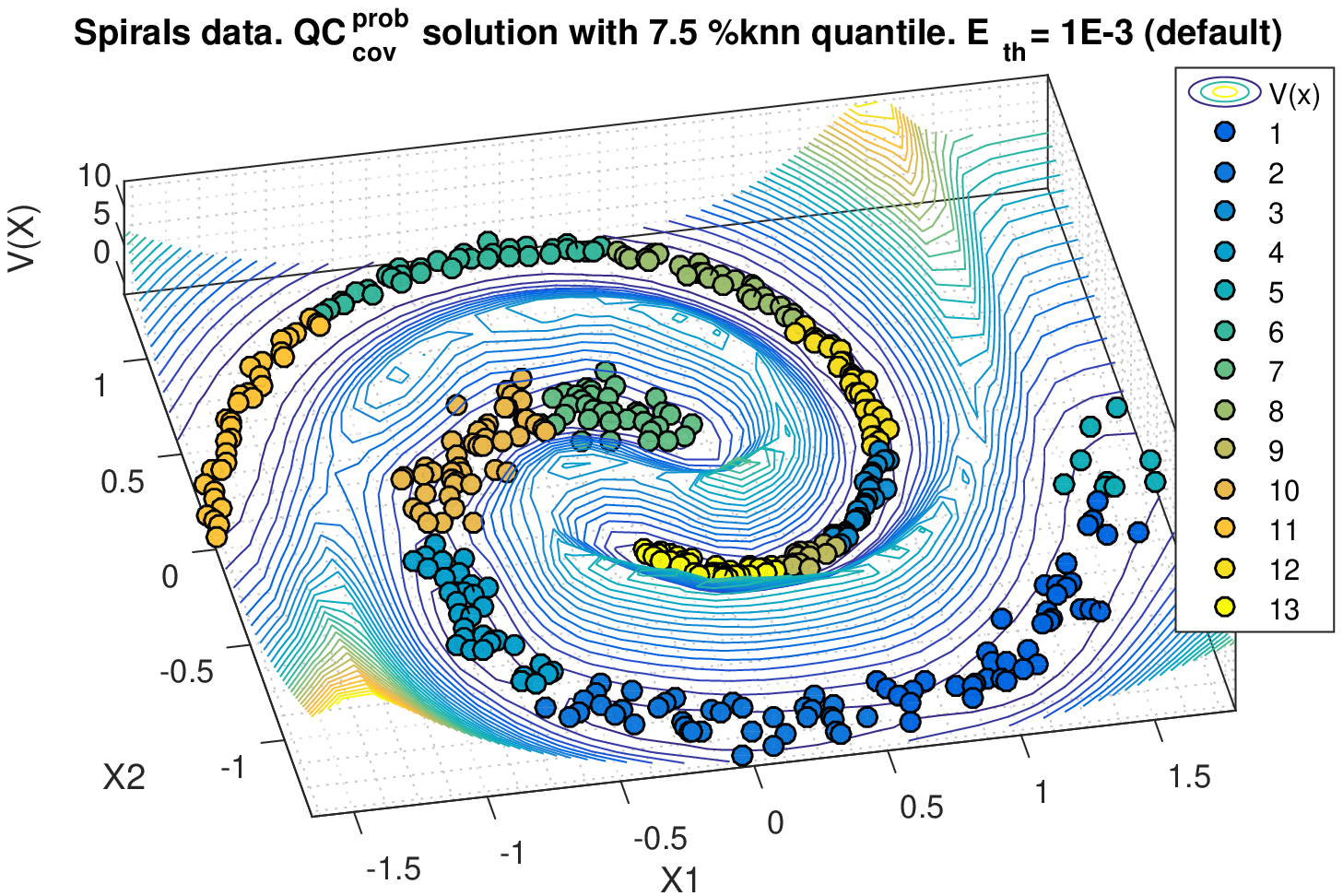}
  \caption{Solution for $QC^{prob}_{knn \, 20\%}$}
  \label{fig:data2_qc3_sol3d_075qtile}
\end{subfigure}%
\begin{subfigure}{.5\textwidth}
  \centering
  \includegraphics[width=0.9\linewidth]{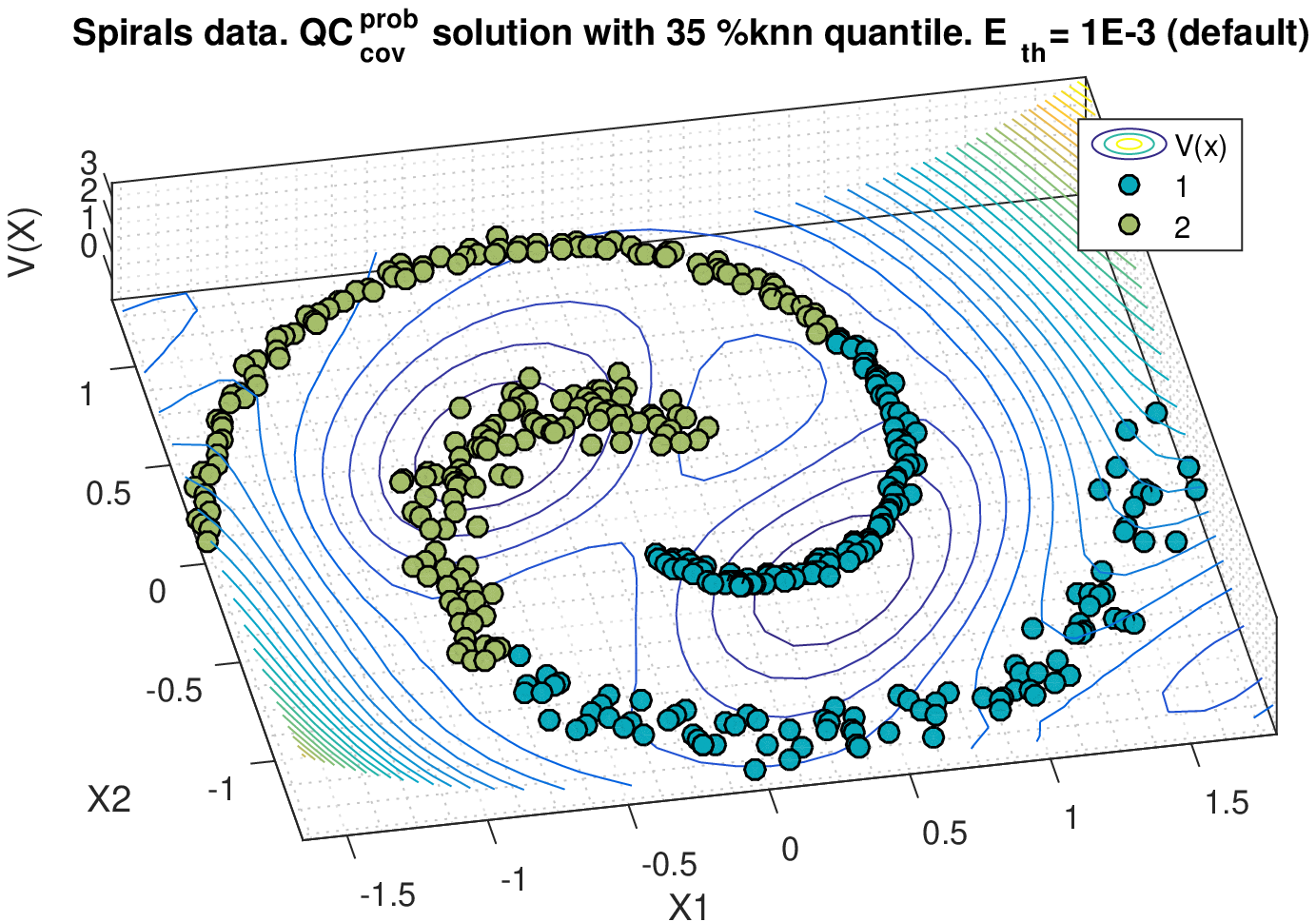}
  \caption{$P(K|X)$}
  \label{fig:data2_qc3_sol3d_35qtile}
\end{subfigure}
\caption{$QC^{prob}_{cov}$ solutions with $E_{th} = 0.001$. The left figure uses a $7.5\%$KNN, here the spirals are not mixed but each one is fragmented in sub-clusters. The right figure uses $35\%$KNN, where the length scale is too big to preserve the spirals not mixed. There are two clusters but the spirals are mixed. These cases show the need of the extended ANLL plots.}
\label{fig:data2_qc3_sol3d}
\end{figure*}

If guided only by the ANLL score in figure~\ref{fig:data2_qc3_knn_js_cv_k}, two local minima would be selected, the first one at $7.5\%KNN$ and the second one at $35\%KNN$, keeping $E_{th}$ with the default value (0.001). Both solutions are illustrated in figure~\ref{fig:data2_qc3_sol3d}.

\begin{figure}[htbp]
\centering
\includegraphics[width=0.9\linewidth]{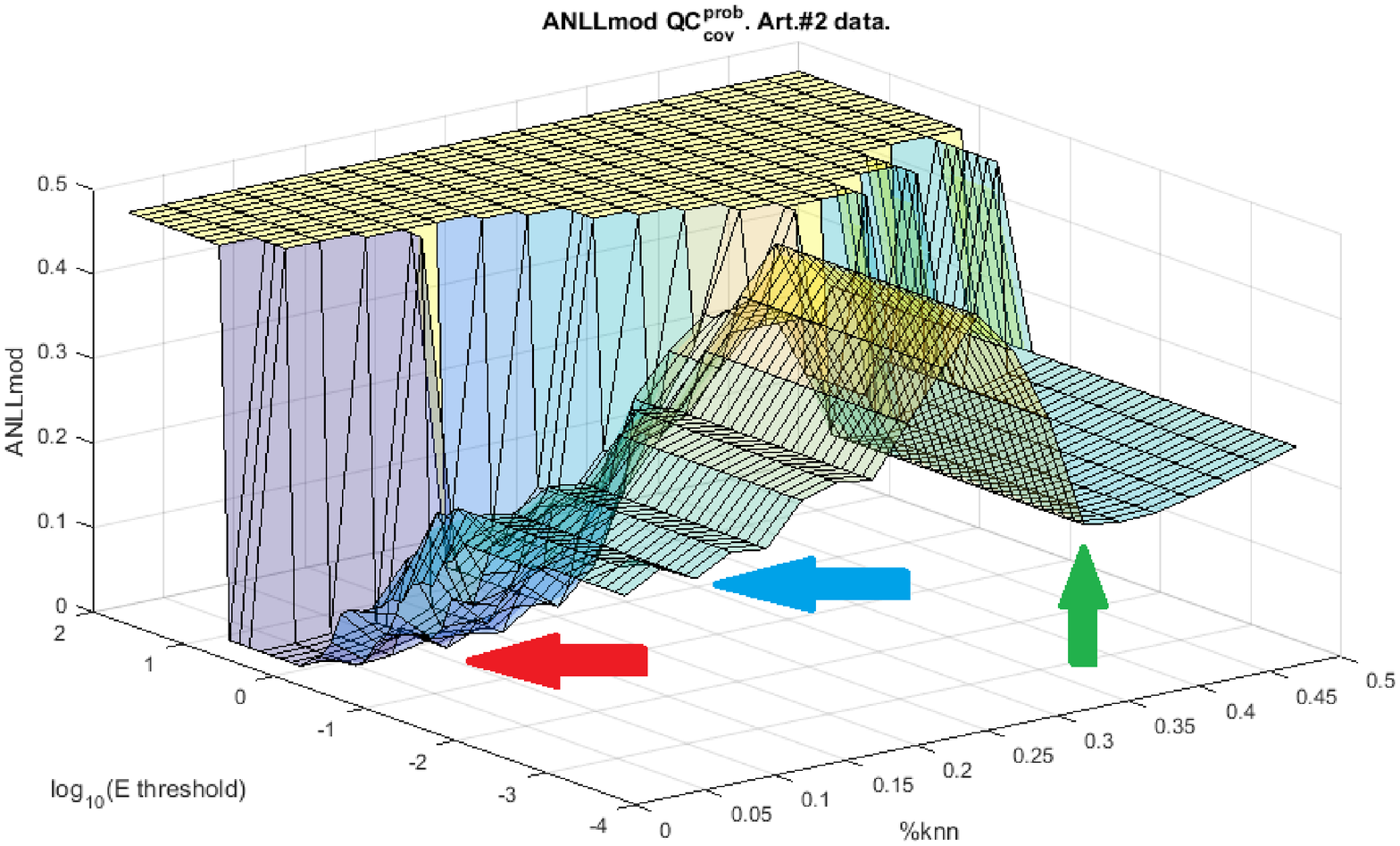}
\caption{Extended ANLL score showing the stability region for high $E_{th}$ values. This region offer a solution based on low length scales where the sub-clusters are merged hierarchically to form the two spirals without being mixed. The ANLL plot indicates three regions of interest: local minima with small length scale (blue arrow), local minima with higher length making a too smooth potential (green arrow), and the stable region of high $E_{th}$ offering the most interesting solution (red arrow).}
\label{fig:data2_qc3_ANLLmod_arrows}
\end{figure}

\begin{figure}[htbp]
\centering
\includegraphics[width=0.9\linewidth]{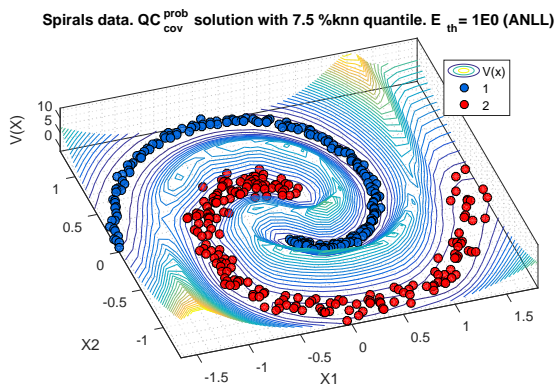}
\caption{Spiral solution based on the stable region parameters in the extended ANLL plot.}
\label{fig:data2_qc3_sol3d_075qtile_Eth1}
\end{figure}

In order to find the optimal solution ($JS =1$), where the spirals are neither mixed nor fragmented, the value of $E_{th}$ should be increased until reaching a region of low ANLL values, as shown in figure~\ref{fig:data2_qc3_ANLLmod_arrows}. The best solution depicted in figure~\ref{fig:data2_qc3_sol3d_075qtile_Eth1}, is achieved in regions with low values of $\%$KNN ($<0.20$) and higher $E_{th}$ ($\in [10^{-1},10^0]$).

Although ANLL is not highly correlated with JS along the $E_{th}$ axis direction, a stable region of low ANLL with high $E_{th}$ implies an underlying hierarchical structure that produces a good JS. Since the JS is not ideally suited for this data set, the expected inverse correlation with ANLL is not present in Table~\ref{table2}.
The stability region varies depending on the QC model, but can be inspected visually using the ANLL plot.

\begin{table*}[!t]
\renewcommand{\arraystretch}{1.3}
\centering
\caption{Data set \#2: Two spirals. In this case, the supervised solution with best JS (only varying $\%KNN$) has a poor performance without modifying the $E_{th}$ parameter. ANLL of the stability region proposes a solution with JS=1.}
\label{table2}
\begin{tabular}{|c|c|c|c|c|c|c|c|c|c|}
\hline
Data \#2 Spirals                          & \textit{Score}                 & $E_{th}$              & \textit{\%KNN}         & $\#K$ & \textit{ANLL}     & \textit{JS}   & $C_v$  & $\rho_{E_{th}}$ & \textit{p-val} \\ \hline
\multirow{4}{*}{$QC_{knn}^{prob}$} & Best JS                   & 0.001            & 47.5          & 1   & 0.510    & 0.50 & -    & 0.60           & 0.005         \\ \cline{2-10} 
                                   & Best ANLL1               & 0.001            & 7.5           & 14  & 0.237    & 0.16 & 1.00    & -               & -             \\ \cline{2-10} 
                                   & Best ANLL2               & 0.001            & 35.0          & 2   & 0.229    & 0.33 & 0.06 & -               & -             \\ \cline{2-10} 
                                   & ANLL  stable at high Eth & $[0.2,0.8]$ & $[2.5,10]$ & 2   & 6.8E-5  & \textbf{1.00} & 1.00    & -               & -             \\ \hline
\multirow{4}{*}{$QC_{cov}^{prob}$} & Best JS                   & 0.001            & 22.5          & 6   & 0.354    & 0.36 & 0.99 & 0.19           & 0.412         \\ \cline{2-10} 
                                   & Best ANLL1               & 0.001            & 7.5           & 13  & 0.223    & 0.17 & 1.00    & -               & -             \\ \cline{2-10} 
                                   & Best ANLL2               & 0.001            & 35.0          & 2   & 0.190    & 0.33 & 0.06 & -               & -             \\ \cline{2-10} 
                                   & ANLL  stable at high Eth & $[0.5,1.5]$  & $[2.5,20]$ & 2   & 1.0E-5 & \textbf{1.00} & 1.00    & -               & -             \\ \hline
\end{tabular}
\end{table*}

TDA methods obtain good performance on the spirals but still lower than $QC^{prob}$, with a $JS=0.79$.

\subsection{Data set \#3: Crabs}
\label{R_crabs}
For the Crabs' data set, ANLL also obtains the appropriate $\%$KNN corresponding with the best JS. Table~\ref{table3} shows that $QC_{knn}^{prob}$ leads to $Js=0.70$ and $QC_{cov}^{prob}$ to $Js=0.74$, respectively. In relation to the extended ANLL score there are no stable hierarchical solutions.

\begin{table*}[!t]
\renewcommand{\arraystretch}{1.3}
\centering
\caption{Data set \#3: Crabs. The supervised solution with best JS matches with the unsupervised solution proposed by ANLL.}
\label{table3}
\begin{tabular}{|c|c|c|c|c|c|c|c|c|c|}
\hline
Data \#3 Crabs                          & \textit{Score}                 & $E_{th}$   & \textit{\%KNN} & \textit{\#K} & \textit{ANLL}  & \textit{JS}   & $C_v$ & $\rho_{E_{th}}$ & \textit{p-val}   \\ \hline
\multirow{3}{*}{$QC_{knn}^{prob}$} & Best JS                   & 0.001 & 17.5  & 4   & 0.110 & 0.74 & 0.90  & -0.83          & 5.7E-6 \\ \cline{2-10} 
                                   & Best ANLL                 & 0.001 & 17.5  & 4   & 0.110 & 0.74 & 0.90  & -               & -       \\ \cline{2-10} 
                                   & ANLL  stable at high Eth & No    & -     & -   & -     & -    & -     & -               & -       \\ \hline
\multirow{3}{*}{$QC_{cov}^{prob}$} & Best JS                   & 0.001 & 15.0  & 4   & 0.126 & 0.70 & 0.89  & -0.88          & 2.8E-7 \\ \cline{2-10} 
                                   & Best ANLL                 & 0.001 & 15.0  & 4   & 0.126 & 0.70 & 0.89  &                 &         \\ \cline{2-10} 
                                   & ANLL  stable at high Eth & No    & -     & -   & -     & -    & -     & -               & -       \\ \hline
\end{tabular}
\end{table*}

\subsection{Data set \#4: Olive oil}
\label{R_olive}

Table\ref{table4} shows the main results for this data set. For the $QC_{knn}^{prob}$, the first ANLL local minimum is closer to the real classification of nine regions but ANLL does not identify the best length scale available: $7.5\%$KNN (JS=0.55) instead of $2.5\%$KNN (JS=0.73). The second ANLL local minimum obtains a similar JS to the best possible one, although the length scale is quite different: $22.5\%$KNN instead of $12.5\%$KNN. Despite not matching exactly with the highest JS, the information provided by the two minima is of paramount relevance, as they point out the two underlying structures, namely, three and nine clusters. The ANLL-JS correlation is quite poor, partly due to ANLL reflects two behaviours but it is compared with two different JS curves.

\begin{table*}[!t]
\renewcommand{\arraystretch}{1.3}
\centering
\caption{Data set \#4: Olive oil. JS in bold refer to the value that should be compared to the corresponding ANNL, depending on whether the model is a solution of the 3-class or 9-class problem. The supervised solutions with best JS match with the unsupervised solutions proposed by ANLL, excepting for the $QC_{knn}^{prob}$ ANLL1 in JS2.}
\label{table4}
\begin{tabular}{|c|c|c|c|c|c|c|c|c|c|}
\hline
Data \#4 Olive                          & \textit{Score}                 & $E_{th}$   & \textit{\%KNN} & \textit{\#K} & \textit{ANLL}  & \textit{JS1}  \textit{JS2}   & $C_V1$  $C_V2$ & $\rho_{E_{th}}$ & \textit{p-val}  \\ \hline
\multirow{5}{*}{$QC_{knn}^{prob}$} & Best JS1 3 regions              & 0.001 & 12.5  & 5   & 0.241 & 0.77 ---   & 0.83  ---     & 0.08            & 7.5E-1 \\ \cline{2-10} 
                                   & Best JS2 9 regions              & 0.001 & 2.5   & 9   & 0.167 & --- 0.73   & --- 0.98           & -0.33           & 1.5E-1 \\ \cline{2-10} 
                                   & Best ANLL1               & 0.001 & 7.5   & 5   & 0.162 & 0.64  \textbf{0.55} & 0.85 0.89      & -               & -      \\ \cline{2-10} 
                                   & Best ANLL2               & 0.001 & 22.5  & 2   & 0.230 &  \textbf{0.74}  0.36  & 0.97  0.95     & -               & -      \\ \cline{2-10} 
                                   & ANLL  stable at high Eth & No    & -     & -   & -     & -          & -              & -               & -      \\ \hline
\multirow{5}{*}{$QC_{cov}^{prob}$} & Best JS 3 regions              & 0.001 & 47.5  & 4   & 0.231 & 0.79   ---   & 0.76   ---  & -0.67           & 1.4E-3 \\ \cline{2-10} 
                                   & Best JS 9 regions              & 0.001 & 20.0  & 8   & 0.187 & ---  0.73    & ---  0.76   & -0.52           & 1.9E-2 \\ \cline{2-10} 
                                   & Best ANLL1               & 0.001 & 15.0  & 9   & 0.175 & 0.52  \textbf{0.72} & 0.99  0.72     & -               & -      \\ \cline{2-10} 
                                   & Best ANLL2               & 0.001 & 45.0  & 4   & 0.220 & \textbf{0.78}  0.41 & 0.76  0.81     & -               & -      \\ \cline{2-10} 
                                   & ANLL  stable at high Eth & No    & -     & -   & -     & -          & -              & -               & -      \\ \hline
\end{tabular}
\end{table*}

Nonetheless, the $QC_{cov}^{prob}$ clearly outperforms $QC_{knn}^{prob}$, ANLL finds solutions with JS practically as good as the best JS ones, the ANLL-JS correlation is better, and the number of clusters is close to the real one (\#K: 4 and 9).

\begin{figure}[htbp]
\centering
\includegraphics[width=0.9\linewidth]{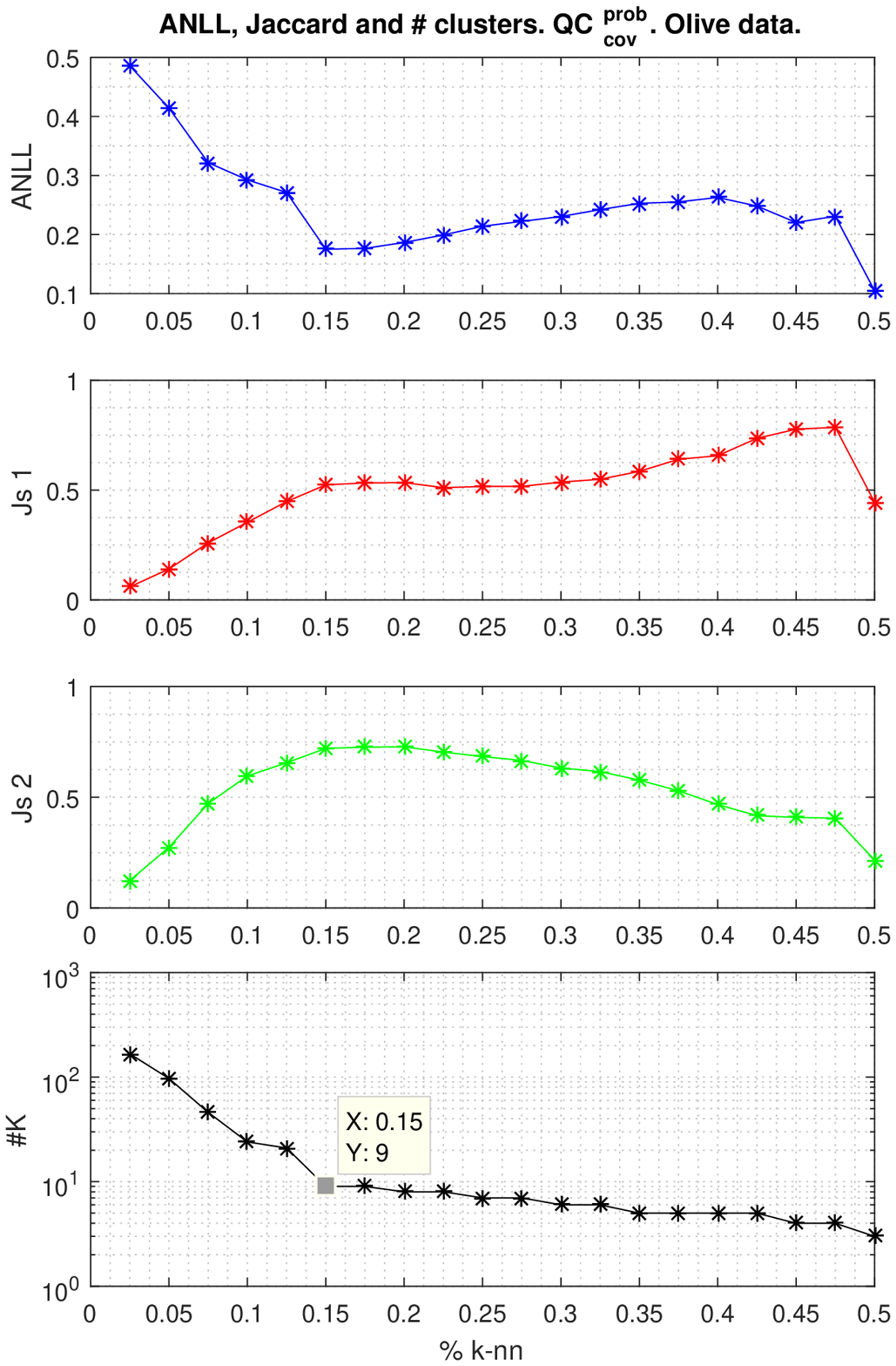}
\caption{ANLL, Jaccard score, $C_v$ and number of clusters obtained by $QC_{cov}^{prob}$ for the olive oil data set. ANLL points out firstly $15\%$ as the most suitable KNNs, and then $45\%$KNN.}
\label{fig:data4_qc3_knn_js_cv_k}
\end{figure}

A further detailed explanation can be obtained observing figure~\ref{fig:data4_qc3_knn_js_cv_k}: The algorithm starts with many sub-clusters with the first KNN; it is important to take into account that dealing with more than 100 clusters is computationally very expensive during the cluster allocation because it has to check many ($100 \cdot 99=9900$) possible paths between potential wells (centroids). Then, the number of clusters decreases drastically until obtaining nine clusters in $15\%$ KNN, and it is here where the first local minimum appears in ANLL, matching with the highest Jaccard score for the structure of nine areas. Then, a subtle local minimum appears at $45\%$ KNN, very close to the highest Jaccard score for the structure of three regions of Italy. Lastly, there is another ANLL minimum at $50\%$ KNN; it is not a real solution but an effect of dealing with very few clusters. The best Jaccard for three regions is $JS=0.73$, and for nine areas is $JS=0.79$.

\section{Conclusion}
\label{Conc}

This paper has presented a novel approach to the detection of the underlying structure in data, within the paradigm of QC. In particular, a merit function to measure goodness-of-fit has been presented in the form of ANLL. This utilises a Bayesian framework to enable optimisation of a control parameter for the estimation of local length scales using set percentages of nearest neighbours. Local minima of  ANLL have empirically shown a high correlation with the highest values of the JS. Therefore, we suggest that ANLL can become a useful objective performance index for unsupervised learning. Furthermore, the ANLL provides useful guidance and insight into QC solutions to detect hierarchical structures in the data.
 
Two new models for PQC with different levels of computational complexity have been proposed. Attending to its simplicity and versatility $QC_{knn}^{prob}$ may outperform $QC_{cov}^{prob}$ in general. However, $QC_{cov}^{prob}$  performs better than $QC_{knn}^{prob}$ when dealing with more challenging data.

The main limitation of $QC_{cov}^{prob}$ stems from its less smooth potential functions as local-covariance kernels have less superposition effect than spherical kernels. As a consequence of this:
\begin{itemize}

\item $QC_{cov}^{prob}$ needs more iterations in the SGD to achieve the same convergence than $QC_{knn}^{prob}$.

\item $QC_{cov}^{prob}$ tends to create more sub-clusters due to the presence of more local minima.
This is not an inconvenience in itself because these sub-clusters can fit better the data and then can be later merged in the cluster allocation process. However, the computation time needed to check all the possible paths between all the centroids may be excessive.
\end{itemize}

The underlying probabilistic framework for QC enables outlier detection as well as the delineation of Bayesian optimal cluster boundaries. QC methods are well-known to have poor performance for high-dimensional data. The proposed framework shares this inherent limitation, the root of which lies in the ultra-metric nature of Euclidean distances in high dimensions as well as sparsity which causes difficulties for local covariance estimation. This remains an area of further work.

\appendices
\label{Appendix}

\section{Improved Cluster Allocation}
\label{app:c_alloc}

Once the gradient descent converges, the observations are allocated to particular potential wells. The first step is to identify the groups of nearby observations that lie in a potential well as a cluster. One of the most robust methods is to apply a community detection algorithm, such as Modularity Maximization~\cite{blondel2008fast}.
To apply the community detection algorithm, the data is transformed into a network using pairwise distances, where the Euclidean distances include the potential values as an extra feature of the data. The adjacency matrix is based on a similarity matrix with Gaussian radial kernel.
Once the communities have been detected, each community being a cluster, we compute the centroids of each cluster by simply averaging their positions.
Now, with the list of clusters and centroids, the goal is to evaluate which is the minimum difference of potential required to cross from \textit{centroid-i} to \textit{centroid-j}. Let us call ``energy'' the minimum difference of potential. The purpose of this step is to merge some nearby sub-clusters that actually belong to the same cluster but were split by the community detection algorithm.

The network is extended with the list of centroids creating a fully connected network, where each node represents an observation or a centroid, and each edge represents the distance between the nodes based on the potential values, $V(\mathbf{x})$.
Using the Dijkstra algorithm~\cite{dijkstra1959note} to find the shortest-path in the network we can build a pairwise measure of the potential differences between centroids (nodes). By way of example, in order to go from the $i$-th node to the $j$-th node, the shortest path found will require at least an ``energy" in terms of the potential units, as:

\begin{eqnarray}
\label{eq:Potdiff1}
\Big. \Delta V(\text{path}) \Big|_{\text{node}_i}^{\text{node}_j} = max\left( V_{\text{path}} \right) - V(\text{node}_i)
\end{eqnarray}

\noindent It should be remarked that the opposite, going from the $j$-th node to the $i$-th node may have a different $\Delta V$.


With this information we can build a pairwise network between nodes with the minimum energy to go from a given node to another one. The next step is to establish a threshold energy to merge any cluster with a similar potential. This procedure is performed to avoid certain situations where several potential wells are connected, for instance by a valley, and hence they have $\Delta V \approx 0$ but not strictly $0$. Therefore, we would consider as the same cluster any pair of potential wells that satisfy:

\begin{equation}
\label{eq:Eth}
\begin{split}
\text{If} \, \big. \Delta V \big|_{\text{node}_j}^{\text{node}_i}  \leq  E_{th} \, \Longrightarrow \, \textit{merge }(\text{cluster}_j, \text{cluster}_i)
\end{split}
\end{equation}

The $E_{th}$ parameter controls the minimum potential difference allowed between two potential wells along their shortest path, $\big. \Delta V \big|_{\text{node}_j}^{\text{node}_i}$; they are considered as different clusters if $\big. \Delta V \big|_{\text{node}_j}^{\text{node}_i}  >  E_{th}$.

If we progressively increase $E_{th}$, more clusters will be merged, up to some point where all the clusters are merged. Thus, this parameter allows the control of the hierarchical structure of the clusters for a specific QC solution defined by $\%$KNN.

\section{Selection of local-covariance threshold}
\label{app:selcovthreshold}

The most appropriate local-covariance threshold for the $QC_{cov}^{prob}$ model is obtained by mapping ANLL (Section~\ref{ANLL}), JS and the number of clusters found ($K$), as a function of the $\%$KNN and the threshold ratio $r$. Finally, the neighbours considered in local-covariance threshold are given by:\par

\begin{center}
\textit{\%K'NN} $ = r $ \textit{(\%KNN)}
\end{center}

\begin{figure}[htbp]
\centering
\includegraphics[width=0.9\linewidth]{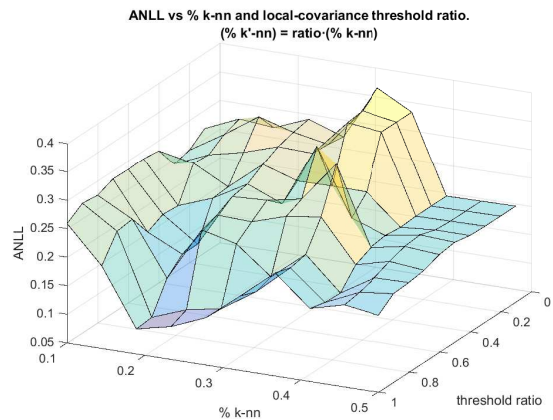}
\caption{ANLL map as a function of $\%$KNN and threshold ratio using ${QC}_{cov}^{prob}$ for artificial data set \#1.}
\label{fig:dat1_anll_local_noise}
\end{figure}

Figure~\ref{fig:dat1_anll_local_noise} shows the ANLL map for the artificial data set \#1. The following conclusions can be drawn:

\begin{itemize}
\item The valley of lowest (best) ANLL values lie around 20$\%$ KNN. However, the most clear valley corresponds with values that range from 0.5 to 1.0; lower ratios lead to a noisy response of ANLL.
\item If $\%$KNN $> 40\%$, only three clusters are found, and it corresponds, in turn, with a biased model, which remains practically invariant to the value of the threshold ratio.
\item The best ANLL values are for ratios from 0.9 to 1.0.
\end{itemize}

\begin{figure}[htbp]
\centering
\includegraphics[width=0.9\linewidth]{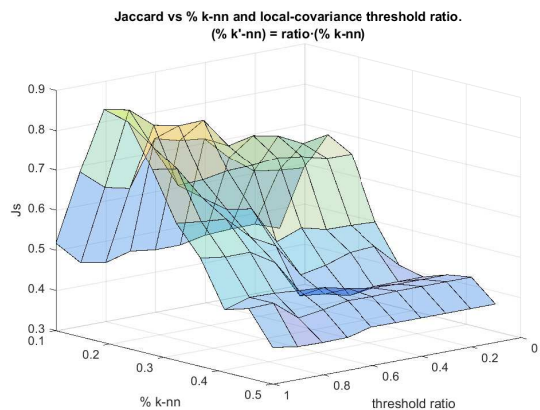}
\caption{Jaccard score map as a function of $\%$KNN and threshold ratio using ${QC}_{cov}^{prob}$ for artificial data set \#1.}
\label{fig:dat1_js_local_noise}
\end{figure}

Figure~\ref{fig:dat1_js_local_noise} shows the corresponding JS map, and also provides useful information:

\begin{itemize}
\item JS is less affected by the threshold ratio than ANLL. It is, though, more sensitive to the \%KNN.
\item The best solutions are located around 20$\%$KNN, as in the case of ANLL.
\item There is a common drop in performance when the ratio decreases; the best values are for $r \, \in \, [0.9, \, 1]$
\end{itemize}

\begin{figure}[htbp]
\centering
\includegraphics[width=0.9\linewidth]{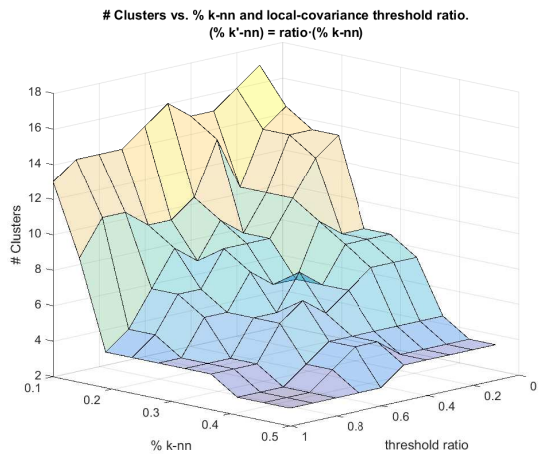}
\caption{Number-of-clusters map as a function of $\%$KNN and threshold ratio using ${QC}_{cov}^{prob}$ for artificial data set \#1.}
\label{fig:dat1_K_local_noise}
\end{figure}

Figure~\ref{fig:dat1_K_local_noise} shows the number-of-clusters map. The main conclusions are, as follows:

\begin{itemize}
\item Both variables, $\%$KNN and threshold ratio are inversely correlated with the number of clusters. This effect was already mentioned in Section~\ref{QC3}; low values of $\%$KNN or $\%$K'NN involve a reduced interaction of each kernel, thus creating an excess of local minima and sub-clusters. The interesting point is to observe that the threshold ratio has a similar influence to $\%$KNN for creating sub-clusters.
\item As the best solutions are for high values of the threshold ratio, one can conclude that an excess of sub-clusters decrease the performance. 
\end{itemize}

In summary, in order to obtain the most simple solution, avoiding spurious sub-clusters, the best threshold ratio is $r \, \in \, [0.9, \, 1]$; for the sake of simplicity, $r=1$ is probably the most sensible choice. That means $\%$K'NN = $\%$KNN.

\section*{Acknowledgments}

This work has been partially supported by the Spanish Ministry of Economy and Competitiveness under project with reference number TIN2014-52033-R supported by European FEDER funds, and by the Spanish Ministry of Education, Culture and Sport under project with reference number PR2015-00217. 


\ifCLASSOPTIONcaptionsoff
  \newpage
\fi



%

%
%

\bibliographystyle{ieeetran}
\bibliography{PhDbib}

%


\begin{IEEEbiographynophoto}{Ra\'ul V. Casa\~na-Eslava}
 obtained Physics Degree in 2009 from University of Valencia (Spain). He spent five years working in industry as technologist and process engineer. Later he returned to the University of Valencia, where in 2015 he obtained a M.Sc. Degree in Electronics. During 2015/ 2016 he has been working as a researcher at Intelligent Data Analysis Laboratory (IDAL) in the University of Valencia (Spain). Nowadays he is a Ph.D. student in the Department of Applied Mathematics, at Liverpool John Moores University (UK). His recent research is focused on Probabilistic Graphical Models, distributed Fisher Information Networks and Quantum Clustering.
\end{IEEEbiographynophoto}

\begin{IEEEbiographynophoto}{Prof. Paulo J. Lisboa}
is Head of Department of Applied Mathematics and Head of the Research Centre for Data Science at Liverpool John Moores University. He has over 250 refereed publications, with awards for citations. His research focus is computer-based decision support in healthcare, sports analytics and computational marketing. In particular, he is interested in rigorous methods to interpret complex models for validation by subject area experts. 
He is co-chair of the Medical Data Analysis Task Force and co-chair of the Big Data Task Force in the Data Mining Technical Committee of the IEEE Computational Intelligence Society and past chair of the Healthcare Technologies Professional Network in the Institution of Engineering \& Technology. He also has editorial and peer review roles in a number of journals and research funding bodies.

\begin{IEEEbiographynophoto}{Dr. Ian H. Jarman}
is a Lecturer in the Department of Applied Mathematics. He is an expert in the analysis of large, high-dimensional data, using both standard statistical techniques and cutting edge non-linear algorithms for knowledge discovery and visualization, and has extensive experience in multi-disciplinary collaborations in: Public Health, Medicine, Retail and Sports Science. He studied Mathematics, Statistics and Computing (1st Class, Honours) at Liverpool John Moores University, and took a PhD in Integrated frameworks for risk profiling of Breast Cancer patients in 2006.

\begin{IEEEbiographynophoto}{Dr. Jos\'e D. Mart\'in-Guerrero}
received a B.S. degree in Physics (1997), a B.S. degree in Electronic Engineering (1999), a M.S. Degree in Electronic Engineering (2001) and a Ph.D. degree in Machine Learning (2004), all from the University of Valencia, Spain. He is currently a Full Professor at the Department of Electronic Engineering, and leader of the Intelligent Data Analysis Laboratory, at the University of Valencia. His research interests include Machine Learning and Computational Intelligence, with special emphasis in Reinforcement Learning and Quantum Machine Learning. He is co-chair and founder Member of the Medical Data Analysis Task Force (Data Mining Technical Committee, IEEE Computational Intelligence Society). He also serves as editor and reviewer for different journals, conferences and research funding bodies.

\end{IEEEbiographynophoto}


\end{IEEEbiographynophoto}


\end{IEEEbiographynophoto}




\end{document}